\documentclass[lettersize,journal]{IEEEtran}
\usepackage{amsfonts}
\usepackage{algorithm}
\usepackage{algpseudocode}
\usepackage{array}
\usepackage[caption=false,font=normalsize,labelfont=sf,textfont=sf]{subfig}
\usepackage{textcomp}
\usepackage{stfloats}
\usepackage{url}
\usepackage{verbatim}
\usepackage{graphicx}
\usepackage{cite}
\usepackage[table,xcdraw]{xcolor}
\usepackage{color}
\usepackage{pifont}
\usepackage{array}
\usepackage{dcolumn}
\usepackage{multirow}

\usepackage{amsfonts}
\usepackage{algorithm}
\usepackage{algpseudocode}
\usepackage{array}
\usepackage[caption=false,font=normalsize,labelfont=sf,textfont=sf]{subfig}
\usepackage{textcomp}
\usepackage{stfloats}
\usepackage{url}
\usepackage{verbatim}
\usepackage{graphicx}
\usepackage{cite}
\usepackage[table,xcdraw]{xcolor}
\usepackage{color}
\usepackage{pifont}
\usepackage{array}
\usepackage{dcolumn}
\usepackage{multirow}

\usepackage{bm}
\usepackage[cmex10]{amsmath}
\usepackage{amssymb}
\usepackage{booktabs}
\DeclareMathOperator*{\argmin}{arg\,min}
\DeclareMathOperator*{\argmax}{arg\,max}

\newcommand\cd[1]{\textcolor{black}{#1}}

\newcommand{\probP}{\text{Pr}}
\newtheorem{assumption}{Assumption}

\usepackage{bm}
\usepackage[cmex10]{amsmath}
\usepackage{amssymb}
\usepackage{booktabs}

\begin{document}

\title{Robust Adaptation of Foundation Models with Black-Box Visual Prompting}

\author{Changdae~Oh, Gyeongdeok~Seo, Geunyoung~Jung, Zhi-Qi~Cheng, Hosik~Choi, Jiyoung~Jung, and~Kyungwoo~Song$^{*}$
        \thanks{C. Oh is with Department of Computer Sciences, University of Wisconsin-Madison, Madison, WI 53706, USA. E-mail: \texttt{changdae@cs.wisc.edu}}
        \thanks{G. Jung and J. Jung are with Department of Artificial Intelligence and H. Choi is with Department of Urban Big Data Convergence, University of Seoul, Seoul, Dongdaemun-gu 02504, South Korea. E-mail: \texttt{\{gyjung975,jyjung,choi.hosik\}@uos.ac.kr}.}
        \thanks{G. Seo and K. Song are with Department of Statistics and Data Science, Yonsei University, Seoul, Seodaemun-gu 03722, South Korea. E-mail: \texttt{\{gd.seo,kyungwoo.song\}@yonsei.ac.kr}}
        \thanks{Z. Cheng is with School of Engineering and Technology, University of Washington, Tacoma, WA 98402, USA. E-mail: \texttt{zhiqics@uw.edu}}
        \thanks{$^*$\textit{Corresponding author}: Kyungwoo Song}
}

\markboth{Journal of \LaTeX\ Class Files,~Vol.~14, No.~8, August~2021}%
{Shell \MakeLowercase{\textit{et al.}}: A Sample Article Using IEEEtran.cls for IEEE Journals}

\maketitle

\begin{abstract}
\cd{
With a surge of large-scale pre-trained models, parameter-efficient transfer learning (PETL) of large models has garnered significant attention. While promising, they commonly rely on two optimistic assumptions: 1) full access to the parameters of a PTM, and 2) sufficient memory capacity to cache all intermediate activations for gradient computation. However, in most real-world applications, PTMs serve as black-box APIs or proprietary software without full parameter accessibility. Besides, it is hard to meet a large memory requirement for modern PTMs. This work proposes black-box visual prompting (BlackVIP), which efficiently adapts the PTMs without knowledge of their architectures or parameters. BlackVIP has two components: 1) Coordinator and 2) simultaneous perturbation stochastic approximation with gradient correction (SPSA-GC). The Coordinator designs input-dependent visual prompts, which allow the target PTM to adapt in the wild. SPSA-GC efficiently estimates the gradient of PTM to update Coordinator. Besides, we introduce a variant, BlackVIP-SE, which significantly reduces the runtime and computational cost of BlackVIP. Extensive experiments on 19 datasets demonstrate that BlackVIPs enable robust adaptation to diverse domains and tasks with minimal memory requirements. We further provide a theoretical analysis on the generalization of visual prompting methods by presenting their connection to the certified robustness of randomized smoothing, and presenting an empirical support for improved robustness.
}
\end{abstract}


\section{Introduction}
\IEEEPARstart{L}{arge}-scale pre-trained models (PTMs), a.k.a. \textit{foundation models} \cite{radford2021learning, achiam2023gpt, liu2023visual}, have driven remarkable advances across diverse domains and applications, based on their outstanding transferability.
By witnessing PTMs' success, Parameter-Efficient Transfer Learning (PETL) methods that efficiently utilize PTMs for downstream tasks have recently emerged. While standard fine-tuning (FT) updates the entire or a large portion of parameters from a PTM, PETL methods aim to achieve a performance comparable to FT by adapting a tiny amount of learnable parameters. 
Among them, \textit{prompt-based approaches} \cite{li2021prefix, jia2022visual, bahng2022visual, khattak2023maple} have been widely investigated on diverse research areas. For PTMs in the vision domain, Visual Prompt Tuning \cite{jia2022visual} is a representative example that injects a few learnable prompt tokens inside of Vision Transformer's \cite{dosovitskiy2020image} layers and only optimizes them. In addition, text encoder-side prompt learning methods for vision-language models are also actively studied \cite{ju2021prompting, zhou2022learning, zhou2022conditional, zang2022unified}. Unlike the above approaches, Bahng et al. \cite{bahng2022visual} initiate an investigation of visual prompting (VP), which attempts to learn a set of parameters in the input RGB pixel space as a \textit{visual prompt} to steer the target PTM.

\begin{figure}[t!]
  \centering
    \includegraphics[width=\linewidth]{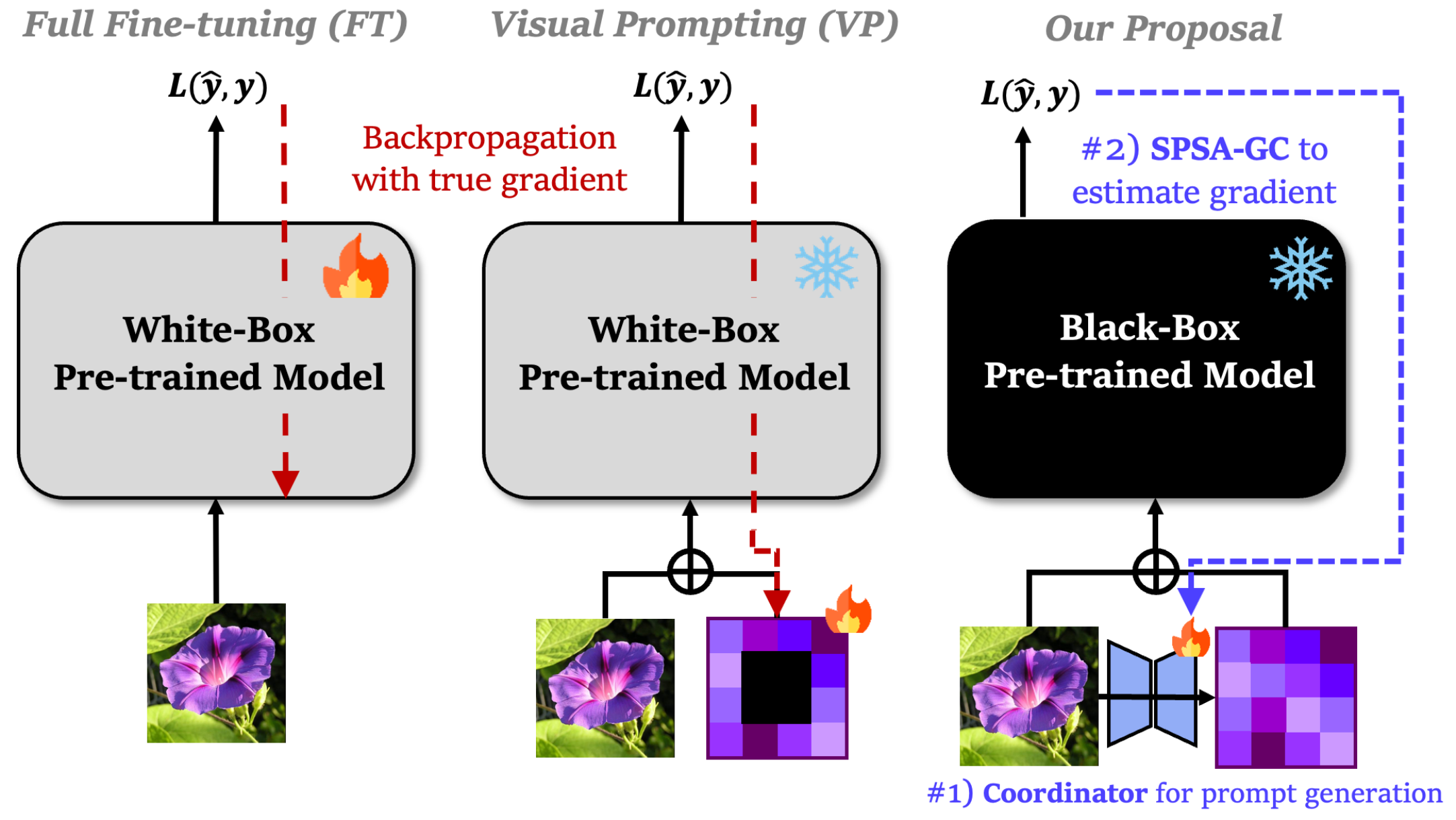}
    \vspace{-1.5em}
    \caption{
    For transfer learning of large-scale pre-trained models (PTM), traditional FT updates the entire model, and VP adapts a few parameters in the input space. However, both FT and VP are only feasible if the PTM's parameters are accessible to the public and require sufficient memory capacity for training. To relax these restrictions, we propose a black-box visual prompting that adopts a zeroth-order optimization (SPSA-GC) to bypass backpropagating a true gradient while reparameterizing the prompt with a tiny learnable network, Coordinator, that produces input-dependent prompts.
    }
\label{fig:overview_illustration} \vspace{-1.7em}
\end{figure}
Although existing PETL methods show impressive potential to adapt PTM with only a few learnable parameters, they rely on two optimistic assumptions that limit their application scope: (1) full PTM parameter accessibility, and (2) a large memory capacity to cache all the intermediate activations for the gradient computation. However, PTMs in real-world applications are commonly served as black-box APIs and proprietary software, and they do not reveal the implementation-level information or full parameters due to commercial issues, e.g., violating model ownership. As a result, \textbf{\textit{exploiting frontier-level PTMs to specific downstream tasks in a black-box setting, where we have limited accessibility to the model's details, is a crucial yet unexplored problem.}} Furthermore, while PETL approaches have a few learnable parameters, they require a large memory capacity to cache intermediate activations to backpropagate the gradient. Therefore, users who want to adopt a large-scale PTM should retain sufficient memory capacity despite the tiny amount of learnable parameters. 

To alleviate the above unrealistic assumptions, we pioneer \textbf{\textit{black-box visual prompting} (BlackVIP)} approach, which enables the parameter-efficient transfer learning of pre-trained black-box vision models from the low-resource user perspective (illustrated in Figure \ref{fig:overview_illustration}). BlackVIP works based on the following two core components: 1) input-dependent visual prompting and 2) a stable zeroth-order optimization algorithm.

First, we augment an input image by attaching a learnable visual prompt per pixel. Note that input space prompting does not require accessibility to model architecture \cite{khattak2023maple, zang2022unified} or the embedding layer weight \cite{zhou2022learning, zhou2022conditional, ju2021prompting}. In contrast to previous work on visual prompting \cite{bahng2022visual} which manually determines a structure of the visual prompt, e.g., a small fraction of a fixed area in an image, we reparameterize the visual prompt with a tiny prompt generation network, \textbf{\textit{Coordinator}}, that produces a corresponding visual prompt per image. As a result, Coordinator automatically designs each prompt based on the input rather than using an input-agnostic manual design. This allows BlackVIP to flexibly change the semantic meanings of an original image. By learning a reparameterized model instead of a prompt itself, we greatly reduce the number of parameters (from 69K of VP \cite{bahng2022visual} to 9K or 1K), so that makes it suitable for the black-box optimization scenario. 

Second, to train the Coordinator, BlackVIP adopts a \textit{zeroth-order optimization} that estimates the gradient from pairs of input queries and outputs of a black-box model. This allows us to adapt any pre-trained vision model, whether its parameters are available in public or not, and can significantly reduce the required memory capacity compared to white-box tuning methods that require caching intermediate activations. For faster and more accurate optimization, we further present a new optimizer, \textbf{\textit{Simultaneous Perturbation Stochastic Approximation with Gradient Correction} (SPSA-GC)} based on (SPSA) \cite{119632}. SPSA-GC first estimates the gradient of the target black-box model based on the output difference of perturbed parameters and then corrects the initial estimates in a momentum-based look-ahead manner. By integrating the Coordinator and SPSA-GC, BlackVIP achieves significant performance improvement over the baseline.

In addition to efficiency, \textit{ensuring robustness of PTMs to distribution shifts} (caused by spurious correlations and noise) is also crucial for safe AI applications. Although previous PETL methods cannot address this directly, we show that BlackVIP can enhance the robustness of PTMs theoretically and empirically, even without any explicit robust training.

This work is an extension of a conference paper \cite{oh2023blackvip}, which has a contribution as a leading work on black-box visual prompting. We summarize our new contributions in this paper as follows: (1) We propose a new variant of BlackVIP, BlackVIP-SE (\textbf{S}tatistical feature-guided \textbf{E}fficient prompt), which is remarkably faster and computationally more efficient than the original BlackVIP while achieving similar classification accuracy with BlackVIP. (2) To investigate the broad applications of our method, we expand the scope of validation with more diverse viewpoints, such as compatibility with text prompting methods, applicability to post-training quantization models, and a white-box transfer regime. (3) For the first time, we provide a theoretical analysis on the generalization mechanisms of visual prompting methods through the lens of the certified robustness of a smoothed classifier.

\section{Related Works}
\label{sec:rel_work}

\subsection{Parameter-Efficient Transfer Learning}
PETL methods pursue fine-tuning of a small subset of PTM's parameter set while achieving competitive performance compared to full FT.
Some of them host new learnable modules inside of PTM \cite{jia2022visual, chen2022adaptformer, chen2023vision} or on top of PTM representation \cite{gao2021clip, zhang2022tip, ouali2023black, zhang2022contrastive}, and others pursue learning tokens of Transformer layer \cite{vaswani2017attention}, e.g., CoOp \cite{zhou2022coop} and its variants \cite{zhou2022conditional, khattak2023maple, zhu2023prompt, ren2024prompt}. Meanwhile, Bahng et al. \cite{bahng2022visual} and its follow-up research \cite{huang2023diversity, chen2023understanding, yang2024fine} explore the Visual Prompting (VP) approach, which introduces a learnable prompt in the input space, neither into the embedding space nor the model's architectural building blocks. The prompt is attached to an image (pixel-wise addition) and updated by gradient descent to optimize the downstream loss function. However, not all existing PETL methods are memory-efficient. That is, they require users to have sufficient memory capacity for backpropagation. More importantly, they assume full accessibility of PTM parameters, which can not be ensured in many real-world cases, where the model is only provided as a black-box API. Although there have been some recent attempts to perform memory-efficient fine-tuning \cite{sung2022lst, liao2023make, kim2024memory}, they still require access to full model parameters. To break these optimistic assumptions, we provide a black-box prompting approach that does not require parameter accessibility or large memory capacity.
\subsection{Black-Box Optimization}
Frontier AI models~\cite{achiam2023gpt} are usually provided through an API or proprietary software to protect ownership. To this end, some recent works fine-tune models via black-box optimization \cite{tsai2020transfer, sun2022bbt, sun2022bbtv2, deng2022rlprompt}. 
These black-box tuning-based works commonly adopt zeroth-order optimization (which exploits estimated gradients) or derivative-free algorithms for parameter updates. For example, BAR \cite{tsai2020transfer} adopts a one-sided approximation gradient estimator. Meanwhile, BBT \cite{sun2022bbt}, and BBTv2 \cite{sun2022bbtv2} adopt CMA-ES \cite{hansen2001completely, hansen2003reducing}, and RLPrompt \cite{deng2022rlprompt} uses reinforcement learning (RL) to optimize the discrete prompts. However, it has been known that derivative-free optimizations (e.g. evolutionary algorithm) are hard to solve large-scale problems and do not guarantee convergence \cite{liu2020primer}. Besides, RL approaches are notorious for their instability and high variance \cite{zhao2011analysis}. This work adopts the Simultaneous Perturbation Stochastic Approximation (SPSA) \cite{119632} as a zeroth-order optimizer, which is known to efficiently provide a gradient approximation on high-dimensional problems \cite{119632, spall1998overview}. Besides, SPSA theoretically guarantees convergence, and the convergence error is linearly upper bounded by the dimension of optimization \cite{119632}. Although SPSA is designed to estimate high-dimensional gradients efficiently, we found that SPSA-based optimization still requires many queries in practice. Thus, we propose SPSA with Gradient Correction (SPSA-GC) that corrects the approximated gradients to enhance the convergence speed. This is the first work exploring the black-box optimization for large PTMs in the vision domain for general-purpose adaptation (rather than a specific domain \cite{tsai2020transfer}).
\begin{figure*}[!t]
    \vspace{-0.2em}
    \centerline{\includegraphics[width=0.875\textwidth]{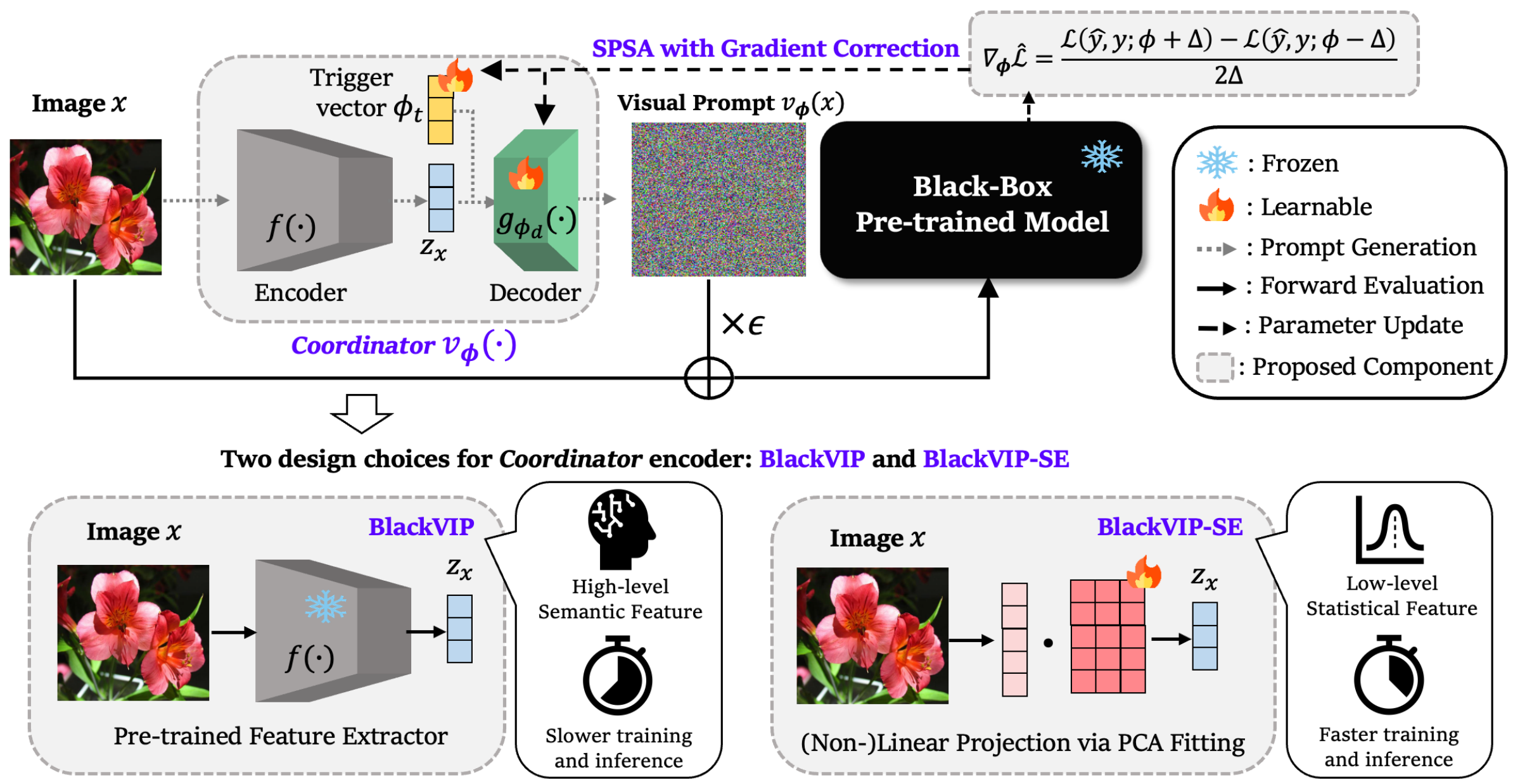}}
    \vspace{-0.4em}
    \caption{We propose an input-dependent prompt designer (Coordinator) and a new zeroth-order optimization algorithm (SPSA-GC) for Coordinator training. Both components are highlighted in the gray-shaded area. The combination of these two proposals forms a novel framework for black-box visual prompting with the following procedure: 1) Coordinator consumes an input image to generate a visual prompt that has the same shape as the image, then 2) the generated prompt is attached to the original image and fed into the black-box target model, finally 3) SPSA-GC takes two outputs induced by differently perturbed Coordinator parameters to estimate the gradient of loss function. Although BlackVIP \cite{oh2023blackvip} adopts another PTM as a frozen feature extractor, this extended work relaxes the reliance on another PTM with a simple projection obtained by a statistical method and names that approach BlackVIP-SE.}
	\label{fig:BlackVIP_framework} \vspace{-1em}
\end{figure*}
\section{Methodology} \label{sec:method}

\noindent\textbf{Background.} Motivated by the remarkable success of the \textit{prompting} paradigm in NLP, Bahng et al. \cite{bahng2022visual} explored a pixel space \textit{visual prompting} (VP) approach for pre-trained vision models. By learning individual RGB pixels as an input space prompt, VP adapts a frozen PTM to targeted downstream tasks without any modification on the model architecture and/or weights. Given the input $x$ and corresponding label $y$, the learning objective of VP is defined as below, 
\begin{align}
	\argmin_{\phi} -\log P(y|x+ \phi) \nonumber
\end{align}
where $\phi$ is a learnable visual prompt and $P(y|x)$ is an output propability of PTM given $x$. During inference, VP employs the shared input-independent prompt for all images. It is noted that $\phi$ is attached to the fixed location, e.g., the outer part of an image like a frame by default. Though AR and VP use different terms and stem from distinct motivations, they share the fundamental concept: \textbf{adapt a PTM to perform new tasks without modifying the model architecture or its weights}. The goal of this work aligns with AR and VP, but we extend and improve them for broader applications.

We introduce a new input-dependent prompt generator, \textit{Coordinator}, with two instantiations for BlackVIP and BlackVIP-SE in Section \ref{sec:method_reparam}. Then, we explain the end-to-end framework with a new zeroth-order optimizer, \textit{SPSA-GC} in Section \ref{sec:method_spsa}. Figure \ref{fig:BlackVIP_framework} shows the overview of our framework.

\subsection{Coordinator: Prompt Reparameterization} \label{sec:method_reparam}
The learning goal of prompting is to minimize the loss function of a downstream task with a frozen PTM by optimizing the input space visual prompt. Given a frozen PTM $P(y|x)$, and prompt-augmented image $\Tilde{x}_{\bm\phi}$ corresponding to the label $y$, the training objective is formulated as:
\begin{align}
\argmin_{\bm\phi} - \log P(y|\Tilde{x}_{\bm\phi}). \nonumber
\end{align}
While VP and AR directly optimize the RGB pixels as learning parameters, we reparameterize the pixel space prompt to a neural network $v_{\bm\phi}(\cdot)$ parameterized by $\bm\phi=\{ \phi_d, \phi_t\}\in \mathbb{R}^{p}$. Specifically, we design a module named \textbf{Coordinator} composed of a fixed encoder $f(\cdot)$, which projects an input image $x$ to a low-dimensional feature $z_{x}$, followed by a lightweight decoder $g_{\phi_d}(\cdot)$ that generates a visual prompt from an integration of instance-specific feature $z_{x}$ and a learnable task-specific \textit{prompt trigger vector} $\phi_{t}$. Consequently, the prompt-augmented image is formulated as follows:
\begin{align}
\Tilde{x}_{\bm\phi} &= \text{clip}(x+\epsilon v_{\bm\phi}(x)), \quad v_{\bm\phi}(x) = g_{\phi_{d}}(z_x, \phi_t)
\label{eq:prompted_image}
\end{align}
\noindent where $z_{x}=f(x)$ is a feature vector of $x$ from the fixed encoder $f(\cdot)$, $\epsilon \in [0,1]$ is a hyperparameter that controls the intensity of visual prompt, and $\text{clip}(\cdot)$ denotes pixel-wise clipping to bound the final image in a valid RGB scale of target model's input transformation \cite{neekhara2022cross}. We concatenate the prompt trigger vector $\phi_{t}$ with $z_{x}$ and then reshape them into a 3D feature map to feed into the decoder $g_{\phi_{d}}(\cdot)$ constructed by a stack of convolution layers. As a result, a rich semantic representation $z_x$ per instance and a task-specific prior $\phi_{t}$ are merged to design a valid visual prompt $v_{\bm\phi}(x)$ for a given image (Supplementary contains full design details). 

Unlike previous works \cite{bahng2022visual, tsai2020transfer} that adopt a pre-defined \textit{input-independent prompt} to a limited region of an image, our approach automatically designs the \textit{input-dependent prompt} that covers an entire image region; thus, it has a higher capability to change the semantics of images if necessary. Thanks to this flexibility, \textbf{it can cover more diverse tasks and be robust to challenging scenarios, e.g., distribution shift.} We demonstrate the efficacy of our prompt design strategies in Section \ref{sec:results}. We now provide two specifications of the encoder component $f(\cdot)$ in the following paragraphs.

\subsubsection{BlackVIP}
We adopt an ImageNet pre-trained model with a self-supervised learning (SSL) objective as a frozen feature encoder $f(\cdot)$ by default. Although the encoder can also be a supervised counterpart or trained from scratch, we choose a frozen SSL encoder for the following three reasons: 1) It has been widely substantiated that self-supervised representation contains multiple discriminative features and spatial information \cite{caron2021emerging, he2022masked, li2021benchmarking, liu2022selfsupervised, huang2022survey, fang2022unleashing, pan2022towards}, so it is more helpful to use a pre-trained SSL encoder than the label-supervised encoder to adapt PTM on various tasks. 2) ImageNet pre-trained models are currently well-publicized \cite{rw2019timm}, so they can be readily adopted by local users, and do not hurt our realistic experimental setting. 3) By leveraging the frozen pre-trained encoder, we significantly reduce the number of learnable parameters compared to the case of training the encoder from scratch. The reduced dimension of optimization contributes to efficient gradient approximation.

\subsubsection{BlackVIP-SE}
While BlackVIP allows us to enjoy the rich features as a source of prompts, it requires an additional forward pass of the frozen encoder. As a result, it increases the runtime during training iterations and inference, compared with model-free prompting baselines, which hinders broader applications. Here, we note an observation that the success of transfer learning of PTMs is not only attributed to high-level feature reuse but also learning from the low-level statistics across data instances \cite{neyshabur2020being}. Motivated by this finding, we hypothesize that low-level statistical features can be an effective alternative to model-based high-level semantic features for producing visual prompts to adapt the black-box target model. To that end, we devise a new variant, BlackVIP-SE (\textbf{S}tatistical feature-guided \textbf{E}fficient prompt), by replacing the auxiliary PTM feature extractor with a more efficient statistical method.
Specifically, we conduct the principal component analysis (PCA) directly on the flattened vectors of training images to find a new coordinate system that preserves the maximum variation of data. Then, we project each image into the PCA-induced low-dimensional space (98 by default) to get statistical features being conveyed to the decoder of the Coordinator\footnote{We first fit the PCA on the few-shot train samples to get a projection matrix and use that matrix as a fixed projector during training and inference.}.
This simple modification endows two favorable merits compared to BlackVIP: 1) it remarkably reduces the runtime (See Table~\ref{tab:runtime}) during training and inference phases by replacing the PTM feature extractor into a single projection matrix, and 2) it also significantly reduces the number of learnable parameters by setting the projection dimension of PCA much lower than that of auxiliary PTM's latent feature dimension in BlackVIP (See Table~\ref{tab:memory}) thereby decreasing the peak memory usage and computational cost.
\begin{algorithm}[t]
\caption{BlackVIP and BlackVIP-SE algorithms}
\label{alg:BlackVIP}
\begin{algorithmic}
    \Require{
    Downstream dataset $\mathcal{D}$, pre-trained model $P(\cdot|\cdot)$, Coordinator $v$ with encoder (or projector) $f$  and prompt decoder $g$, is parameterized by $\bm\phi_{i}=\{\phi_{d,i}, \phi_{t,i}\}$, SPSA-GC decaying parameters $\{ a_{i},c_{i} \}$, smoothness parameters $\beta$, prompt intensity $\epsilon$, and training iteration $T$.}
    \State \textcolor{gray}{// Initialize $\bm\phi_{1}=\{\phi_{d,1}, \phi_{t,1}\}$, $\{a_{1}, c_{1}\}$ and $m_{1}$ }
    \For{$i$ in $1$ to $T$}
    \State \textcolor{gray}{// Parse a batch $(x, y) \sim \mathcal{D}$ and design the prompt}
    \State $v_{\bm\phi_{i}}(x) = g_{\phi_{d,i}}(f(x), \phi_{t,i})$
    \State $\Tilde{x}_{\bm\phi} = \text{clip}(x+\epsilon v_{\bm\phi_{i}}(x))$
    \State \textcolor{gray}{// Draw a sample $\Delta_{i}$, set $c_{i}$, and estimate the gradient}
    \State $L(\bm\phi_{i}) :=  - \log P_{\bm\phi_{i}}(y|\Tilde{x}_{\bm\phi})$
    \vspace{0.35em}
    \State $\hat{e}_{i}(\bm\phi_{i}) = \frac{L(\bm\phi_{i} + c_i\Delta_{i}) - L(\bm\phi_{i} - c_i\Delta_{i})}{2c_{i}\Delta_{i}}$
    \vspace{0.2em}
    \State \textcolor{gray}{// Set $a_{i}$, and update parameters}
    \State $m_{i+1} = \beta m_{i} - a_{i}\hat{e}_{i}(\bm\phi_{i} + \beta m_{i})$
    \State $\bm\phi_{i+1} = \bm\phi_{i} + m_{i+1} $
    \EndFor 
\end{algorithmic}
\end{algorithm}
\subsection{End-to-End Black-Box Visual Prompting}
\label{sec:method_spsa}
Unlike other PETL approaches, we consider the PTM as a black-box predictor that gives only a prediction output (i.e. post-softmax probability vector) for a given input image query. In this black-box setting, we adopt the zeroth-order optimizer, SPSA, with our deliberate modification to optimize Coordinator without requiring the oracle true gradient.
\subsubsection{SPSA} Spall et al. proposed Simultaneous Perturbation Stochastic Approximation (SPSA) \cite{119632, SPALL1997109} that approximates the high-dimensional gradient efficiently. Given the positive decaying sequences of $a_{i}>0$ and $c_{i}\in [0,1]$, the gradient estimate, $\hat{e}$, and single-step parameter update of SPSA is described as follows:

\vspace{-1em}
{\small
\begin{align}
\hat{e}_{i}(\bm\phi_{i}) &= \frac{L(\bm\phi_{i} + c_i\Delta_{i}) - L(\bm\phi_{i} - c_i\Delta_{i})}{2 c_i} \Delta_{i}^{-1}  \\
\bm\phi_{i+1} &= \bm\phi_{i} - a_i \hat{e}_{i}(\bm\phi_{i})
\label{eq:spsa}
\end{align}
}
\noindent where $L$ is an objective, $\bm\phi_{i}\in \mathbb{R}^{p}$ is $p$-dimensional learnable parameters, and $\Delta_{i} \in \mathbb{R}^p$ is a $i^{th}$-step random perturbation vector, sampled from zero-mean distributions that satisfy finite inverse moments condition \cite{119632, spall2003} such as Rademacher and Segmented Uniform distribution. With only two forward evaluations, i.e., calling twice the model API, SPSA estimates the gradient from the difference between model outputs of two different perturbations, and we can use this estimated gradient to update the parameters $\bm\phi$ of Coordinator.

\subsubsection{Gradient Correction} Although the standard SPSA works well in myriad applications \cite{4469948, stein13_interspeech, 9143831}, it may suffer slow convergence in practice \cite{880982, Spall1997AcceleratedSS}, and this convergence speed issue gets even worse on high-dimensional settings such as neural network optimization. We speculate that the source of slow convergence is its noisy gradient estimation caused by the irrelevant direction of randomly sampled perturbations or intrinsic data noise. To mitigate the estimation noise, inspired by Nesterov's accelerated gradient (NAG) \cite{Nesterov1983AMF}, we improve the parameter update rule in Eq. \ref{eq:spsa} as below:
\begin{align}
\bm\phi_{i+1} &= \bm\phi_{i} + m_{i+1} \\ 
m_{i+1} &= \beta m_{i} - a_{i}\hat{e}_{i}(\bm\phi_{i} + \beta m_{i}) 
\label{eq:spsa_gc}
\end{align}
\noindent where $\beta \in [0,1)$ is a smoothing parameter. As clearly noted in \cite{pmlr-v28-sutskever13}, when the poor update, which increases the target loss, occurs, this NAG style update strongly pulls it back towards the original parameter $\bm\phi_{i}$. For example, SPSA-GC computes the direction of parameter update at position $\bm\phi_{i} + \beta m_{i}$ rather than $\bm\phi_{i}$, and obtains the momentum-based update vector $\beta m_{i}- a_{i}\hat{e}_{i}(\bm\phi_{i} + \beta m_{i})$. Here, while the momentum makes the gradient estimates smoother, Nesterov-style lookahead mechanism enables us conservative gradient computation per step--results in ``gradient correction". This correction effect mitigates the high volatility optimization trajectory filled with noisy gradient estimations, which is problematic in many cases of large-scale black-box optimization \cite{ghadimi2013stochastic, mashkaria2023generative} by progressing through the prediction-correction loop. Moreover, the NAG-style update rule excels in the larger learning rate optimization regime \cite{pmlr-v28-sutskever13}, which is standard practice for the VP optimization problem \cite{bahng2022visual}. Given the strongly stochastic nature of SPSA and the large learning rate practice of VP, we expect that this \textit{gradient correction} can be highly effective in reducing variance in estimation \cite{park2025zip} and increasing its convergence.
\section{Results} \label{sec:results}

\begin{table}[thb]
\centering
\small
\caption{Results on synthetic datasets: correlation shifts and varying object location scenarios.} 
\vspace{-0.6em}
\begin{tabular}[!t]{@{}l|cc|cc@{}}
\toprule
& \multicolumn{2}{c}{Biased MNIST} & \multicolumn{2}{c}{Loc-MNIST} \\ 
 & $\rho=0.8$ & $\rho=0.9$ & 1:1 & 1:4 \\ \cmidrule(r){1-5} 
\cellcolor{lightgray}{VP (white-box)} & \cellcolor{lightgray}{57.92} & \cellcolor{lightgray}{43.55} & \cellcolor{lightgray}{86.79} & \cellcolor{lightgray}{86.54} \\ \cmidrule(r){1-5}
ZS & 37.56 & 37.25  & 29.70 & 22.70 \\
BAR & 53.25 & 53.07 & 33.98 & 26.05 \\
VP w/ SPSA-GC & 60.34 & 53.86  & 16.21 & 25.68  \\
BlackVIP-SE & 55.73 & 55.93  & 62.92 & 55.57  \\
BlackVIP & \textbf{66.21} & \textbf{62.47} &  \textbf{69.08} & \textbf{60.86} \\
\bottomrule
\end{tabular} \vspace{-1.5em}
\label{tab:syn_results} 
\end{table} 
\subsection{Experimental Setup} \label{sec:exp_setup}
\noindent{\textbf{Datasets.}} To investigate the importance of prompt design, we consider two synthetic datasets that simulate the correlation shift and varying object location scenarios (see Section \ref{sec:toy} and Supplementary). Then, we extensively evaluate our methods on 14 transfer learning benchmarks. These cover diverse visual domains and tasks, that require understanding various visual semantics like scenes, actions, fine-grained categories, textures, satellite imagery, the number of objects, and natural objects. Moreover, to evaluate the robustness of prompting PTM in the wild, we additionally consider $L_{2}$-noise perturbation scenario (in Figure \ref{fig:thm_illu}) and three benchmark datasets of distribution shifts from WILDS \cite{koh2021wilds} in Supplementary.

\begin{figure}[h!]
     \centering
        \includegraphics[width=0.485\linewidth]{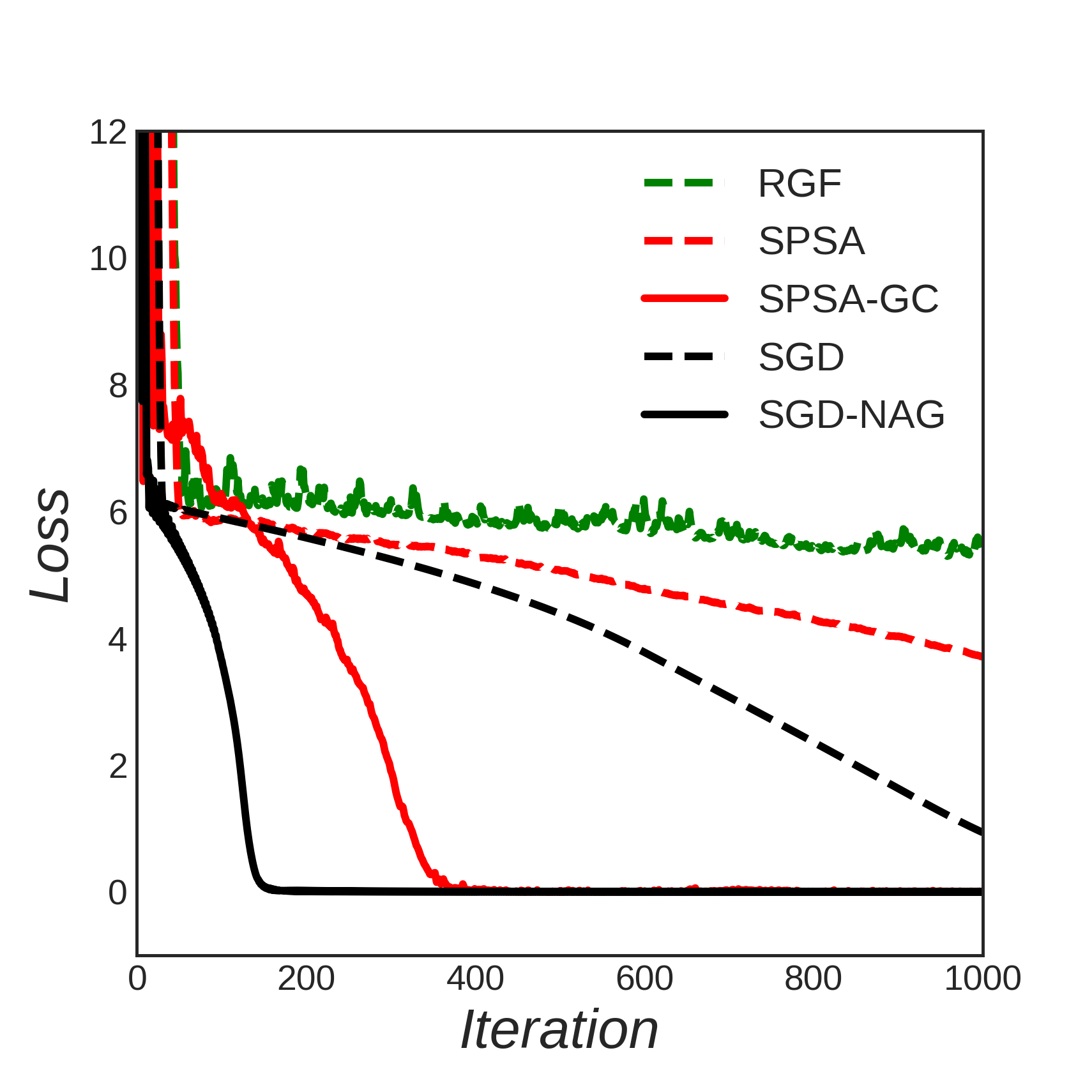}
        \vspace{-0.5em}
        \includegraphics[width=0.485\linewidth]{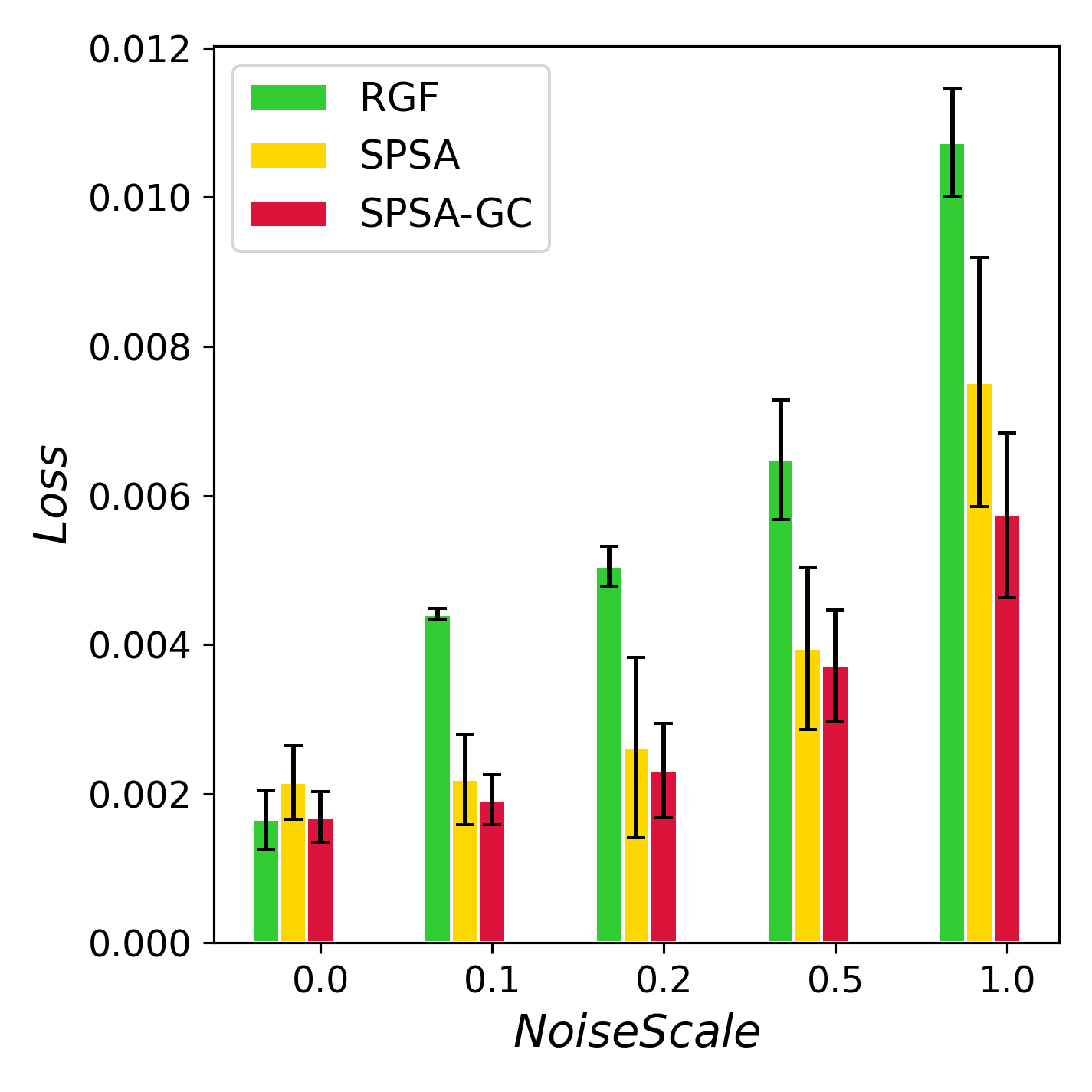}
        \vspace{-0.5em}
    \caption{(Left) loss curve and (right) noise sensitivity analysis of 100-Dimensional Rosenbrock optimization problem with ZOO algorithms.}
\label{fig:opt}
\vspace{-0.7em}
\end{figure}
\noindent{\textbf{Backbone model and methods.}} We mainly adopt CLIP ViT-B/16 \cite{radford2021learning} as a target PTM because CLIP does not require a separate classifier for different tasks and performs classification dynamically with text embedding features by text-side prompting (Supplementary Table IV provides results on ViT-L/14 as well). As the encoder part of Coordinator, we use ImageNet pre-trained \texttt{vit-mae-base} checkpoint for BlackVIP and sci-kit-learn's vanilla PCA or kernel PCA for BlackVIP-SE. 
As baselines, we consider 1) zero-shot classification (ZS) with the standard text prompt ``\texttt{a photo of \{classname\}}", 2) black-box adversarial reprogramming (BAR) \cite{tsai2020transfer} that embed the downsized images of the downstream task inside a learnable prompt, and 3) VP with SPSA-GC that replaces the pure gradient descent in VP \cite{bahng2022visual} with our SPSA-GC. 
Following the few-shot classification setup of \cite{zhou2022coop}, we use 16 samples per class for training in all evaluations by default. More details are provided in Supplementary.

\subsection{Experiments on Synthetic Datasets} \label{sec:toy}
\subsubsection{Comparison between optimizers}
We validate our SPSA-GC on the well-known optimization benchmark, the Rosenbrock function. We report the normalized loss ($\frac{|L(\theta^{*})-L(\theta)|}{|L(\theta^{*})-L(\theta_{0})|}$) where $L(\theta^{*})$ and $L(\theta_{0})$ is the loss value on the optimal and initial point, respectively, and $L(\theta)$ is a loss value on the current parameter $\theta \in \mathbb{R}^{100}$. In Fig. \ref{fig:opt} (left), SPSA-GC shows faster and more stable convergence than Random Gradient-Free (RGF) adopted in previous works \cite{liu2018zeroth, tsai2020transfer}, and even achieves a comparable result to Nesterov's Accelerated Gradient (SGD-NAG) method, which adopts the true gradient. Besides, we simulate the noisy loss observation scenario (emulating the mini-batch optimization) by adding Gaussian noise to the loss, i.e., $L_{noisy}(\theta)=L(\theta)+\xi$, where $\xi \sim \mathcal{N}(0,\sigma^{2})$. As $\sigma$ increases, RGF rapidly degenerates while SPSA is still relatively stable, and our gradient correction (SPSA-GC) gives further improvement. This verifies the robust gradient approximation of SPSA-GC. Beyond the small-scale optimization problem, we compare multiple black-box optimizers, including evolutionary algorithms, reinforcement learning methods, and other zeroth-order methods, in Fig.~\ref{fig:bbo-extended} of Section~\ref{sec:few}

\subsubsection{Robustness to Correlation Shift}
Next, we evaluate our method on Biased MNIST \cite{bahng2020learning} to investigate the robustness of BlackVIPs' input-dependent automatic prompt design under the feature-label correlation shift environment. Biased MNIST is a modified version of MNIST \cite{lecun1998gradient}, constructed to validate a model's generalization ability under color bias shift. At the train-time, each digit has a unique preassigned background color that strongly correlates with the target label. The degree of correlation is determined by the value $\rho \in [0, 1]$, and the correlation ratio is reversed as $1-\rho$ at the inference-time. Results are summarized in Tab. \ref{tab:syn_results}. In this setup, BlackVIPs remarkably outperform others (even white-box VP), and the performance gap becomes larger under the stronger correlation. This means that BlackVIPs, by flexibly modifying the semantics of the image with input-dependent visual prompts, can be beneficial in challenging adaptation scenarios wherein spurious correlation exists.

\subsubsection{Robustness to Varying Object Location}
For many object recognition datasets, the target objects are commonly placed in the center of the image. However, input data of foundation models in the wild may have varying object locations. We expect that BlackVIP and BlackVIP-SE adopt input-dependent prompts that cover the entire region of the image so they are still robust even if the object is not always located in the center of the image. To validate this, we create a variant of the MNIST, Loc-MNIST, by putting a real target digit on the four edges and an arbitrary fake digit in the center of the black blank image. The location of the target digit and the class of the fake digit are chosen randomly. We further consider a more challenging setup in that the fake digit is four times larger (1:4) than the real one. We summarize the results in Tab. \ref{tab:syn_results}. Compared to the manually designed input-independent prompts (BAR and VP) that are frame-shaped by default, conditional prompts produced by BlackVIPs achieve significantly better performance, which supports the superiority of the Coordinator's prompt design.

\subsection{Few-shot Transfer Learning on Benchmarks}\label{sec:few}
\begin{table*}
\vspace{-0.1em}
\caption{Top-1 classification accuracy on 14 benchmarks that require natural, specialized, structured, and fine-grained visual recognition. BlackVIP and BlackVIP-SE show outstanding results among input-space visual prompting methods. \textit{Win} means the number of datasets that each method beats the zero-shot performance. Gray-colored values are the results of white-box prompt learning with stochastic gradient descent. All experiments are done in 16-shot training with three repeated runs.}
\vspace{-0.9em}
\label{tab:fs14}
\begin{center}
\begin{small}
\resizebox{\textwidth}{!}{\begin{tabular}{@{}l|cccccccccccccc|c|c@{}}
\toprule
Method & Caltech & Pets  & Cars & Flowers & Food & Aircraft & SUN & DTD & SVHN & EuroSAT & RESISC & CLEVR & UCF & IN & \textit{Avg.} & \textit{Win} \\ \midrule
\cellcolor{lightgray}{VP (white-box)}  & \cellcolor{lightgray}{ 94.2} & \cellcolor{lightgray}{ 90.2} & \cellcolor{lightgray}{ 66.9} & \cellcolor{lightgray}{ 86.9} & \cellcolor{lightgray}{81.8} & \cellcolor{lightgray}{ 31.8} & \cellcolor{lightgray}{ 67.1} & \cellcolor{lightgray}{ 61.9} & \cellcolor{lightgray}{60.4} & \cellcolor{lightgray}{ 90.8} & \cellcolor{lightgray}{81.4} & \cellcolor{lightgray}{40.8} & \cellcolor{lightgray}{ 74.2} & \cellcolor{lightgray}{ 67.4 } & \cellcolor{lightgray}{ 71.1} & \cellcolor{lightgray}{13}\\ \midrule
ZS & 92.9 & 89.1 & 65.2 & 71.3 & 86.1 & 24.8 & 62.6 & 44.7 & 18.1 & 47.9 & 57.8 & 14.5 & 66.8 & 66.7 & 57.6 & - \\
BAR & \textbf{93.8} & 88.6 & 63.0 & 71.2 & 84.5 & 24.5 & 62.4 & \textbf{47.0} & 34.9 & \textbf{77.2} & \textbf{65.3} & 18.7 & 64.2 & 64.6 & 61.4 & 6 \\
VP w/ SPSA-GC & 89.4 & 87.1 & 56.6 & 67.0 & 80.4 & 23.8 & 61.2 & 44.5 & 29.3 & 70.9 & 61.3 & 25.8 & 64.6 & 62.3 & 58.8 & 4 \\
BlackVIP-SE  & 93.5 & \textbf{89.8} & \textbf{65.6} & \textbf{71.5} & 86.3 & \textbf{25.4} & \textbf{65.3} & 45.1 & \textbf{46.1} & 71.2 & 61.1 & 32.1 & 68.5 & \textbf{67.1} & 63.5 & \textbf{14} \\ 
BlackVIP  & 93.7 & 89.7 & \textbf{65.6} & 70.6 & \textbf{86.6} & 25.0 & 64.7 & 45.2 &  44.3 & 73.1 & 64.5 & \textbf{36.8} & \textbf{69.1} & \textbf{67.1} & \textbf{64.0} & 13 \\  \bottomrule
\end{tabular}}
\end{small}
\end{center}
\vspace{-0.15em}
\end{table*}
\subsubsection{Main Results}
\begin{figure}[htbp]
    \footnotesize
     \centering
     \includegraphics[width=0.75\linewidth]{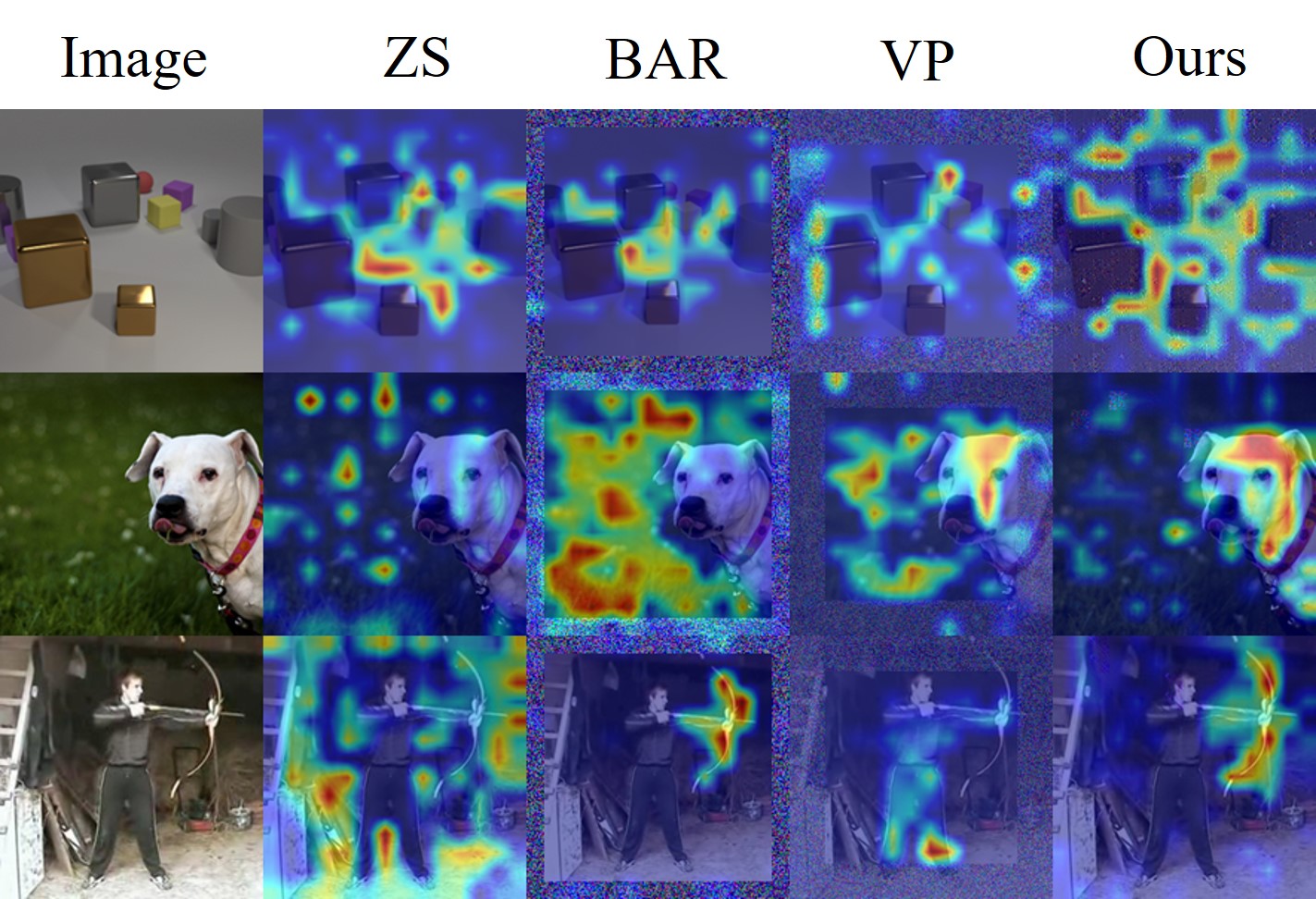}
     \vspace{-0.25em}
    \caption{Grad-CAM analysis on CLEVR, Pets, and UCF101.}
    \label{fig:gcam_main}
    \vspace{-0.5em}
\end{figure}

We conduct a main validation on 14 commonly used benchmarks \cite{zhou2022coop, zhou2022conditional, bahng2022visual}. As shown in Tab. \ref{tab:fs14}, while BAR and VP undergo large performance variations across datasets, BlackVIPs show consistently high performance, i.e., improve performance from ZS on 13 over 14 datasets for BlackVIP and 14 over 14 for BlackVIP-SE. Specifically, BAR shows promising results on tasks that require understanding coarse semantics (DTD \cite{6909856}, EuroSAT \cite{helber2019eurosat}, and RESISC \cite{7891544}), but does not show competitiveness in CLEVR \cite{johnson2017clevr} that requires visual reasoning (counting objects) by capturing the general semantics of the image. Meanwhile, BlackVIP performs well on various tasks by extending or limiting the region of attention of the black-box PTM (Fig. \ref{fig:gcam_main}), which denotes that BlackVIP is a high-capability prompt learner that adapts a PTM to various downstream tasks. Overall, BlackVIP-SE also shows remarkable performance gains that are comparable to BlackVIP but significantly faster.

\begin{figure}[htbp]
    \footnotesize
     \centering
     \vspace{-0.25em}
     \includegraphics[width=0.7\linewidth]{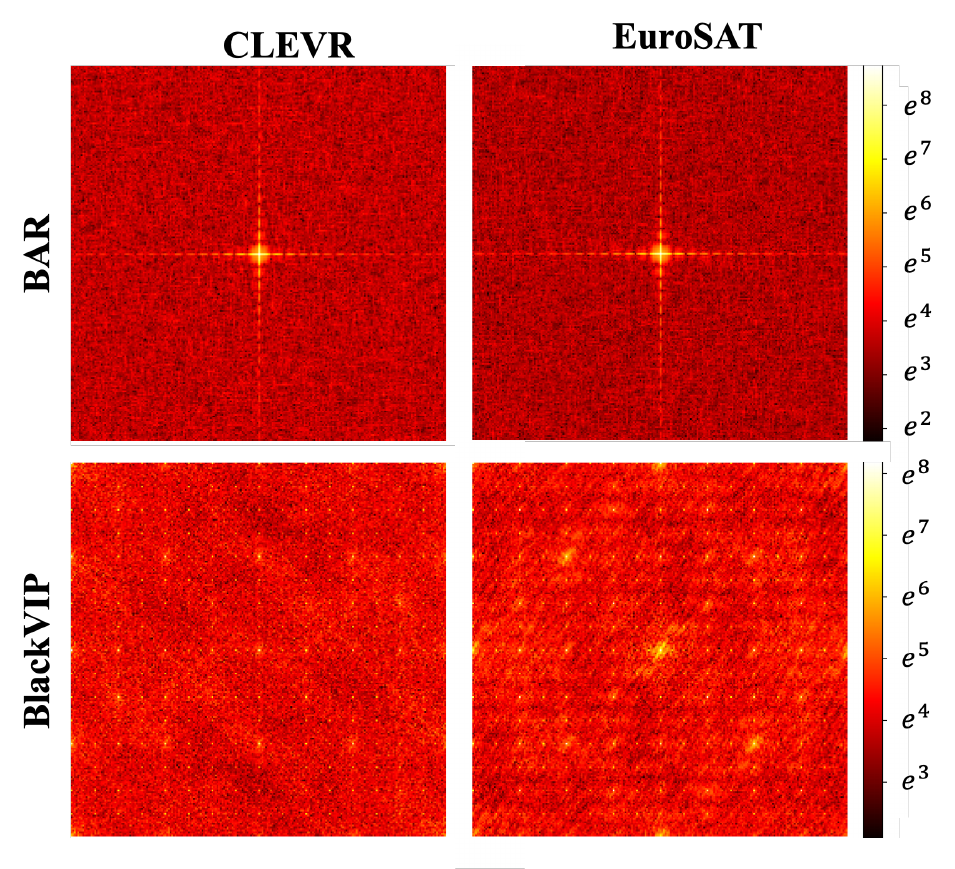}
     \vspace{-0.75em}
    \caption{Frequency domain analysis on learned prompt visualized in log-scale Fast Fourier transform average features.}
    \vspace{-0.65em}
    \label{fig:freq_anal}
\end{figure}

We speculate that BlackVIP-SE excels in data sets in which the input feature space has a relatively low intrinsic dimensionality \cite{pope2021intrinsic}, such as SVHN, so PCA-based approaches can find a sufficiently discriminative low-dimensional manifold \cite{hastie2009elements}. Meanwhile, BlackVIPs slightly underperform BAR on tasks that require scene-level understanding, which indicates that prompts attached to the entire image somewhat hinder the coarse recognition of scene-level semantics. We investigate this further through a frequency-domain analysis by applying the Fast Fourier transform~\cite{brigham1988fast,tsao2024autovp} on the learned visual prompt samples of BAR and BlackVIP. As we can see in Fig.~\ref{fig:freq_anal}, which depicts the average frequency features across 128 random prompt samples, BAR learns the prompts highly concentrated in the low-frequency domain, whereas BlackVIP induces prompts of mid/high-frequency domains. This implies that BlackVIP's input-dependent global prompts can excel at fine-grained recognition tasks, whereas existing local input-agnostic prompts can be preferred for some coarse-grained tasks such as scene understanding. 

\subsubsection{Practical Usefulness}
\begin{table}[h]
\centering
\caption{Train-time peak memory allocation and the number of learnable parameters on ImageNet-1K with batch size 64.}
\scriptsize
\begin{tabular}{@{}lcccc@{}}
\toprule
\multirow{2}{*}{Method} & \multicolumn{2}{c}{Peak Memory (MB)} & \multicolumn{2}{c}{Params} \\ \cmidrule(l){2-5}
 & \multicolumn{1}{c}{ViT-B} & \multicolumn{1}{c}{ViT-L} & ViT-B & ViT-L \\ \midrule
FT (white-box) & 21,655 & 76,635 & 86M  & 304M  \\ 
LP (white-box) & 1,587 & 3,294 & 513K & 769K  \\
VP (white-box) & 11,937 & 44,560 & 69K & 69K \\
\midrule
BAR & 1,649 & 3,352 & 37K  & 37K \\
VP w/ SPSA-GC & 1,665 & 3,369 & 69K & 69K  \\
BlackVIP-SE & \textbf{1,319}  &\textbf{ 2,540 }&\textbf{ 1K} & \textbf{1K} \\ 
BlackVIP & 2,428 & 3,260 & 9K & 9K \\ \bottomrule
\end{tabular} \label{tab:memory}
\vspace{-0.4em}
\end{table}

\noindent{\textbf{Memory and computation efficiency (Table \ref{tab:memory}). }}First, BlackVIP and BlackVIP-SE have 9K and 1K parameters in total, respectively. This allows us to remarkably reduce the computational cost compared with other white-box approaches (such as FT and LP) or black-box approaches (BAR and VP). Furthermore, all the black-box adaptation approaches significantly reduce the train-time peak memory requirement, which is burdening for white-box approaches. Among the black-box methods, BlackVIP-SE achieves the best memory-and-computation efficiency.

\begin{figure}[htbp]
\centering
\vspace{-0.15em}
\includegraphics[width=\linewidth]{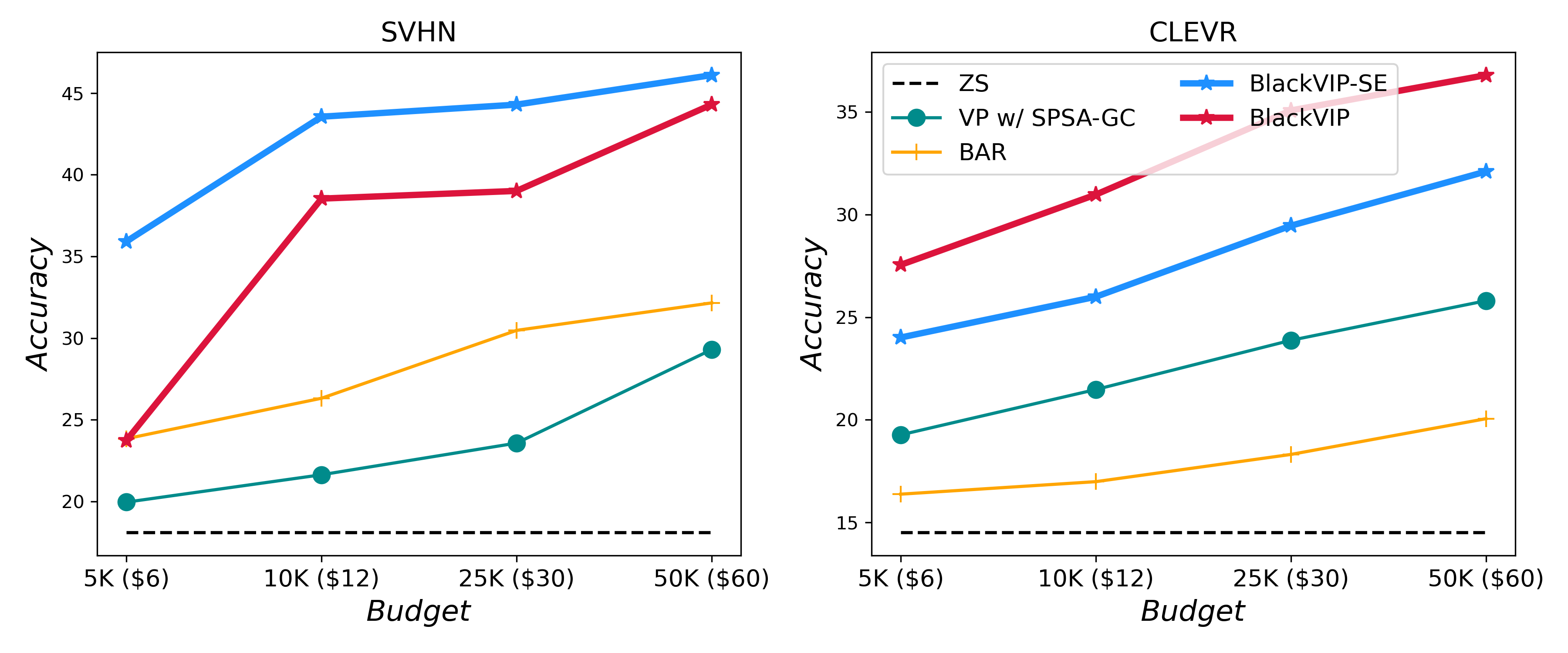}
\vspace{-0.75em}
\caption{Query efficiency. (x-axis) The number of API calls for forward evaluation and the cost for achieving (y-axis) corresponding performance.}
\vspace{-0.5em}
\label{fig:query_efficiency}
\end{figure}

\noindent{\textbf{Query efficiency (Figure \ref{fig:query_efficiency}). }}Second, BlackVIPs show outstanding query efficiency among black-box prompting methods. For instance, by sending just 10K API calls with 12 USD (based on Clarifai Vision API), we can improve the performance of a zero-shot model by about two or three times better using BlackVIP and BlackVIP-SE. Given that we pay for each query to adapt black-box API models, ensuring query efficiency does matter, and our BlackVIPs offer a favorable cost-and-performance trade-off. However, BlackVIP is still par less-efficient than optimization with ground-truth gradient, which requires 50 times fewer iterations to achieve the same final performance, highlighting the necessity of following up study to improve query efficiency further.

\begin{table}[t]
    \centering
    \caption{Runtime (second) comparison between prompting methods. Training denotes the sum of the forward prompt generation and the parameter update procedures for a single batch, and Inference indicates per-batch forward time.}
    \vspace{-0.2em}
    \footnotesize
    \begin{tabular}{@{}lccccc@{}} \toprule 
    Runtime & ZS     & BAR    & VP     & BlackVIP-SE & BlackVIP \\ \midrule 
    Training  & -      & 0.1402 & 0.1417 & 0.1233      & 0.2684   \\ 
    Inference & 0.0044 & 0.0049 & 0.0049 & 0.0053      & 0.0098   \\ \bottomrule 
    \end{tabular} \label{tab:runtime}
    \vspace{-0.6em}
\end{table}

\noindent{\textbf{Runtime analysis (Table \ref{tab:runtime}).}} Next, we investigate the wall-clock time for the single-batch forward prompt generation and parameter update during training and inference time per batch. While BlackVIP achieves superior classification accuracy on many setups, it compromises the time complexity due to its reliance on an auxiliary feature extractor during prompt generation. This hinders the broad use of BlackVIP in some applications where the inference time is crucial. In contrast, our new variant, BlackVIP-SE, addresses this issue by replacing the feature extractor model with a single projection matrix founded by PCA. As a result, BlackVIP-SE achieves a competitive runtime compared with other baselines (BAR and VP) while significantly outperforming in terms of classification accuracy (See Table \ref{tab:fs14}). This reduced runtime during training and inference phases allows it to become a favorable alternative to BlackVIP in time-critical applications.

\begin{figure}
    \centering
    \includegraphics[width=0.85\linewidth]{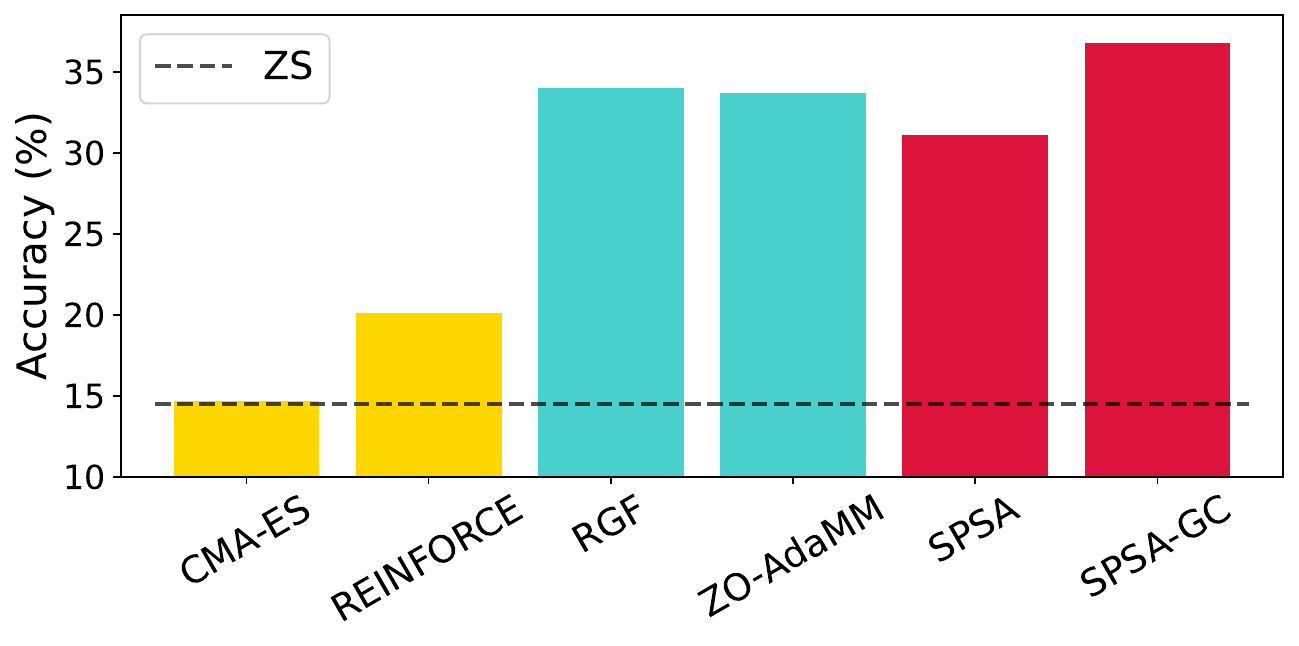}
    \vspace{-1em}
    \caption{Comparison between different optimization methods for BlackVIP Coordinator training on CLEVR dataset.}
    \label{fig:bbo-extended}
    \vspace{-0.5em}
\end{figure}

\noindent{\textbf{Comparison between optimizers (Figure \ref{fig:bbo-extended}).}} We compare SPSA-GC with several representative alternatives from evolutionary algorithms, reinforcement learning, and zeroth-order methods, including CMA-ES~\cite{hansen2001completely}, REINFORCE~\cite{williams1992simple}, RGF~\cite{liu2018zeroth}, and ZO-AdaMM~\cite{chen2019zo} for BlackVIP Coordinator training on the CLEVR dataset. While other zeroth-order methods, such as RGF and ZO-AdaMM, beat the vanilla SPSA, they underperform SPSA-GC, demonstrating the effectiveness of our proposal in a large-scale optimization setup.

\begin{table}[h]
\centering
\caption{Combination of different black-box visual and text prompts with a frozen target model, where CuPL and LLM BBO denote large language model-based optimized prompts.}
\scriptsize
\begin{tabular}{@{}llcc@{}}
\toprule
Visual Prompt      & Text Prompt               & EuroSAT & DTD   \\ \midrule
-          & CuPL \cite{pratt2023does}  & 45.91   & 55.85 \\
BAR         & CuPL \cite{pratt2023does}  & 55.22   & 55.42 \\
BlackVIP-SE & CuPL \cite{pratt2023does}  & 64.34   & \textbf{55.91} \\
BlackVIP    & CuPL \cite{pratt2023does}  & \textbf{68.88}   & 55.58 \\ \midrule
-          & LLM BBO \cite{liu2024language} & 52.80    & 45.15 \\ 
BAR         & LLM BBO \cite{liu2024language} & 59.85   & 44.58 \\
BlackVIP-SE & LLM BBO \cite{liu2024language} & 70.05   & 45.41 \\
BlackVIP    & LLM BBO \cite{liu2024language} & \textbf{71.80}    & \textbf{45.59} \\ \bottomrule
\end{tabular} \label{tab:bbtextprompt}
\vspace{-0.5em}
\end{table}
\subsection{Compatibility with black-box text prompting} Although this work focuses on prompt optimization for visual inputs, it has also been actively studied to optimize text prompts under the black-box target model assumption. As stated in the white-box setting \cite{zang2022unified}, prompts for each modality differently affects the embedding space with complementary benefits. We investigate the compatibility of black-box visual prompting and two black-box text prompt optimization methods \cite{pratt2023does, liu2024language}. Tab.~\ref{tab:bbtextprompt} shows that our methods show strong compatibility with those text prompting methods overall.

\subsection{Further Analysis: BlackVIP v.s. BlackVIP-SE}
\begin{figure}[htbp]
     \centering
     \vspace{-0.15em}
     \includegraphics[width=0.67\linewidth]{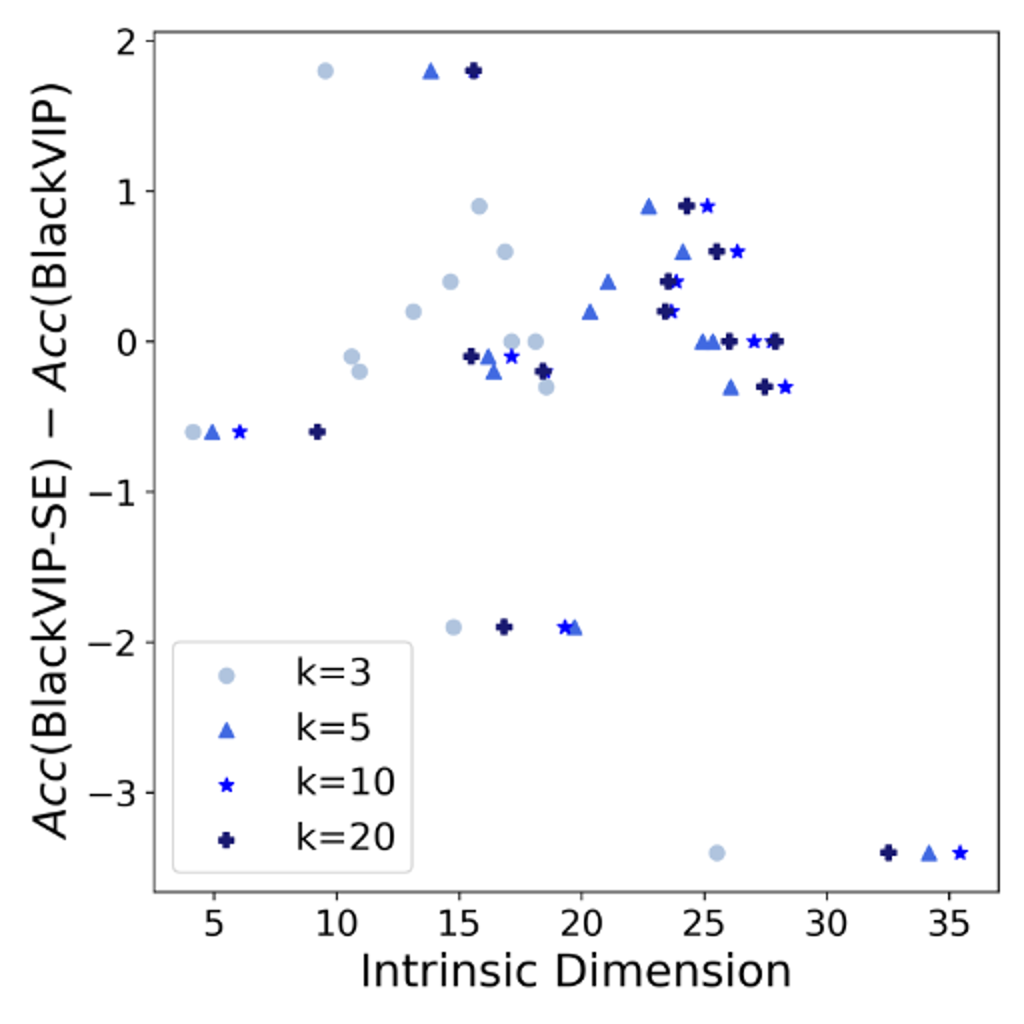}
     \vspace{-0.75em}
    \caption{Relationship between intrinsic dimensionality estimates (with varying nearest neighbor parameter K of 
 the estimator) and relative performance between BlackVIP and BlackVIP-SE across different datasets.}
    \label{fig:intrinsic}
    \vspace{-0.65em}
\end{figure}

It is surprising to observe that PCA-based simple low-dimensional features can rival the high-dimensional features of a pre-trained visual encoder. We hypothesize that the effectiveness of BlackVIP-SE depends on \textit{intrinsic dimensions} of datasets \cite{pope2021intrinsic}, that is, as the intrinsic dimension of a dataset is smaller, PCA features can be more effective in producing a meaningful visual prompt, resulting in rivaling performance with features from a pre-trained visual encoder. To verify this, we estimate the intrinsic dimensions in 13 benchmark datasets using the maximum likelihood estimation method proposed in \cite{NIPS2004_74934548, mackay_ghahramani_2005} and see the relationship between the relative performance of BlackVIP-SE compared to BlackVIP. As shown in Figure \ref{fig:intrinsic}, the intrinsic dimension of the dataset is negatively correlated with the relative performance of BlackVIP-SE, which means that the effectiveness of low-level statistic features depends on datasets' intrinsic dimension. 

However, one should be careful to adopt BlackVIP-SE when there exist spurious features--exist in training samples while lacking in test samples--in the target dataset. To be specific, although BlackVIP-SE achieves comparable performance to BlackVIP across all real-world benchmarks, it significantly underperforms BlackVIP on synthetic datasets, such as Biased MNIST in Table~\ref{tab:syn_results}, that have explicit spurious features.

\subsection{Application on White-box Transfer Learning} \label{sec:broad_app}
\begin{table}[!ht]
\scriptsize
    \centering
    \caption{White-box transfer learning with true gradients access. Gray rows denote methods that host a large number of parameters.}
    \begin{tabular}{@{}clc|cccc@{}}
    \toprule
        Mod. & Method & Params & SVHN & CLEVR & Pets & \textit{Avg.} \\ \midrule
        - & ZS & - & 18.1 & 14.5 & 89.1 & 40.6 \\
        - & FT & 86M & 71.5 & 36.2 & 93.9 & 67.2 \\ \midrule
        L $\&$ V & MaPLe & 1.2M & 66.5 & 40.5 & 92.9 & 66.6 \\ 
        L $\&$ V & LoRA & 150K & 45.1 & 27.3 & 89.0 & 53.8 \\ \midrule
        L & CoOp & 9K & 43.8 & 36.1 & 91.7 & 57.2 \\
        L & CoCoOp & 35K & 48.6 & 29.7 & 93.0 & 57.1 \\ 
        L & LoRA & 150K & 42.6 & 36.5 & 89.7 & 56.3 \\ \midrule
        V & VP & 69K & 60.4 & 40.8 & 90.2 & 63.8 \\
        V & VPT & 73K & 62.9 & 39.9 & 93.2 & 65.3 \\
        V & LoRA & 150K & 45.9 & 25.7 & 91.0 & 54.2 \\
        V & Ours-SE & 1K & 60.5 & 36.1 & 90.0 & 62.2 \\
        V & Ours & 9K & 60.4 & 42.6 & 90.8 & 64.6 \\
        V & Ours & 68K & 61.8 & 43.6 & 90.5 & 65.3 \\ 
        V & Ours & 150K & 61.4 & 45.4 & 91.4 & 66.1 \\ \bottomrule
    \end{tabular} \label{tab:whitebox}
    \vspace{-0.65em}
\end{table}
While the core motivation of BlackVIP was from the desire to break the parameter accessibility constraint and overwhelming memory requirements, our prompt reparameterization strategy not only resolves the memory issues but also improves the design of visual prompts compared to other visual prompting approaches. To validate the efficacy of the prompt designs of BlackVIPs, we replace the estimated gradients of optimizers with a true gradient computed by the first-order optimizer, assuming we have access to the entire model parameters. As shown in Table \ref{tab:whitebox}, our prompting approaches (corresponding to BlackVIP and BlackVIP-SE) outperform other recent prompt-based efficient tuning methods such as VP, VPT, LoRA, CoOp, and CoCoOp under similar parameter capacity. Even our approach (with 150K parameters) is comparable with MaPLe \cite{khattak2023maple}, which requires eight times more parameters, implying that our innovative design of the prompt can also be adopted in white-box transfer regimes.

\section{Theoretical Understanding on the Generalization of the Visual Prompting} \label{sec:theory}

\begin{figure*}[t]
\centering
\includegraphics[width=\textwidth]{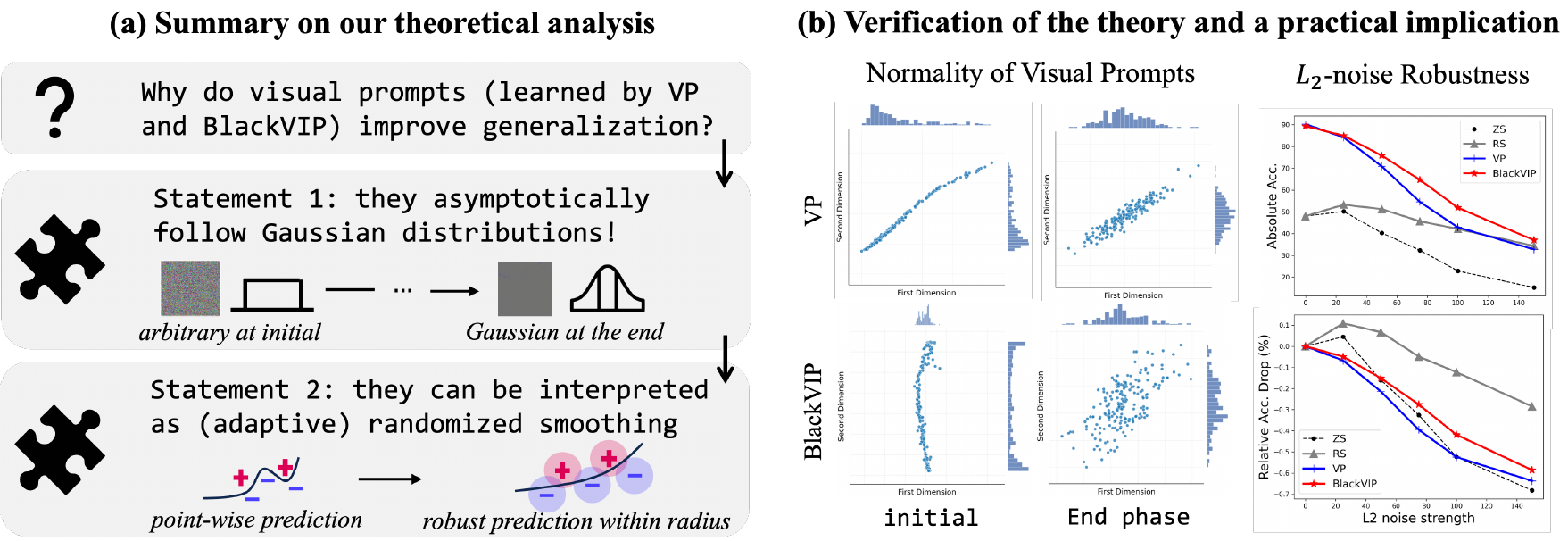} \vspace{-1.75em}
\caption{(a) Summary of our theoretical analysis, and (b) empirical verification of the theory with a practical implication. All experiments are conducted in a white-box setup using true gradient to clearly demystify the generalization behavior of visual prompting. We plot empirical distributions for the first two dimensions of visual prompts at the initial and end phases of training to confirm their asymptotic normality at the left of (b). Under this normality and other mild assumptions, VP and BlackVIP act as an adaptive RS classifier, thereby ensure robustness to $L_2$-noise to some extent as shown in the right side of (b).}
\label{fig:thm_illu}  \vspace{-1.5em}
\end{figure*}

In this section, we provide a new hypothesis to explain why visual prompting methods such as VP and BlackVIP can improve the generalization of the pre-trained target models. Specifically, we will first show that visual prompts learned by VP and BlackVIP asymptotically follow Gaussian distributions. Given that result, we reveal that pre-trained classifiers enhanced by visual prompts are equivalent to classifiers augmented by randomized smoothing, thereby guarantee generalization under $L_2$ feature shifts and robustness against $L_{2}$ noise (See Supplementary Material A for full derivation).

\noindent\textbf{Preliminary.} A classifier $\mathcal{F}(\cdot)$ maps a $d$ dimensional input $x\in \mathbb{R}^d$ to a categorical variable $c\in\mathcal{C}$. We define a randomized smoothing (RS) classifier $\mathcal{G}(x):=\argmax_{c} \probP(\mathcal{F}(x+\zeta)=c)$ that returns a class $c$ that maximizes an aggregated classification probability under Gaussian noise $\zeta \sim \mathcal{N}(0,\sigma^2 \mathbf{I})$ with an intensity $\sigma$ and an identity matrix $\mathbf{I}$. Recent works \cite{cohen2019certified, hong2022certified} showed that a smoothed classifier maintains its prediction against $L_{2}$-norm bounded perturbations, where the robustness guarantee of the smoothed classifier originated from the additive noise following a Gaussian distribution. Here, we note that the optimization trajectory of a visual prompt by SGD with a constant learning rate can be viewed as a sampling process from a Gaussian posterior centered at the local optimum \cite{stephan2017stochastic, wu2021revisiting}. This indicates a prompted classifier imitates the behavior of a smoothed classifier, ensuring robustness against input perturbations at test time. The Gaussian approximation requires Assumptions 1-3 specified in Supplementary, which allow us to approximate the discrete-time SGD dynamics by a continuous-time stochastic process. 
As shown in Figure \ref{fig:thm_illu}, where we visualize the first two dimensions of the visual prompts, although they are not Gaussian in the initial training phase, they end up approximating Gaussian distributions, showing the practicality of our assumption. While we assume SGD updates for simplicity, we demonstrate in Supplementary A that the Gaussian approximation of parameters remains valid for black-box settings, specifically SPSA and SPSA-GC.

\subsection{Visual Prompting as Adaptive Randomized Smoothing} 
Recall the form of BlackVIP loss function: $L(\phi) = -\log{P(y|x+v_\phi(x))},$
where $P(y|x)$ is the PTM output probability for ground truth $y$ given $x$ and $v_\phi(x)$ is a visual prompt parameterized by $\phi$. This expression is reduced to a learning objective of VP, when $v_\phi(x)=\phi$. A prompt from BlackVIP depends on data $x$. We optimize the visual prompt $\phi$ via SGD with a constant learning rate $\alpha$ as $\phi_{t+1}=\phi_t-\alpha \nabla L(\phi_t)$. As shown in a previous work \cite{stephan2017stochastic}, given a sufficiently large time step $t$, when the $\phi_t$ converges to the minimum of loss function and $\nabla L(\phi_t)$ approaches zero, $\phi^{*}=\phi_{t}=\phi_{t+1}, ...$ becomes samples from a stationary Gaussian as below:

\vspace{-0.5em}
{\small
\begin{equation}
    \phi^* \sim N(\mu_{\phi^*}, \Sigma_{\phi^*}), \quad \Sigma_{\phi^*}\mathbf{A}+\mathbf{A}\Sigma_{\phi^*}=\frac{\alpha}{S}\mathbf{B}\mathbf{B}^T, \label{eq:vp_bayesian}
\end{equation}}
where $S$ is the minibatch size. We view $\phi^*$ as a random variable near the minimum of the loss function, showing Gaussian approximation after the time step $t$ from initialization. Here, $\mu_{\phi^*}:=\argmin_{\phi} -\log{P(y|x+v_\phi(x))}$ is the point where our loss function is minimized, and $\mathbf{A}$ is a Hessian of loss w.r.t. the optimal point of $\phi^*$ and $\mathbf{B}\mathbf{B}^T$ is a covariance matrix of the SGD's gradient noise (See \cite{stephan2017stochastic}). Now, given that we repeatedly encounter the same sample data instance $x$ several times during training, we get the following statement,

\vspace{-1em}
{\small
\begin{align*}
\mathcal{G}_{\phi^*}(x)& =\argmax_{c\in \mathcal{C}} \probP(\mathcal{F}(x+v_{\phi^*}(x))=c)\\
(i) \,\; v_{\phi^*}(x)&=\phi^* \sim \mathcal{N}(\mu_{\phi^*}, \Sigma_{\phi^*}) \quad \textrm{(VP)}\\
(ii) \, v_{\phi^*}(x)& \approx v_{\mu^*}(x)+J_{\mu^*}(x)(\phi^*-\mu_{\phi^*}) \\ &\sim \mathcal{N}(v_{\mu^*}(x), J_{\mu^*}(x) \Sigma_{\phi^*} J_{\mu^*}(x)^\top) \quad \textrm{(BlackVIP)}
\label{eq_}
\end{align*}}
\noindent where $J_{\mu^*}(x)=\partial v_{\phi^*}(x)/\partial\phi^* \vert_{\phi^*=\mu_{\phi^*}}$ and $v_{\mu^*}(x):=v_{\phi^*}(x)\vert_{\phi^*=\mu_{\phi^*}}$. This statement reveals that a classifier with visual prompting can be interpreted as a randomized smoothing classifier $\mathcal{G}_{\phi^*}(x)$ where the location parameter ($\mu$) of the Gaussian distribution is adapted to a direction in which the downstream classification loss should be minimized. This allows us to leverage the robustness guarantee \cite{cohen2019certified, hong2022certified} of a smoothing to explain the generalization of visual prompting.

\subsection{Certified Robustness of Visual Prompting Classifier}
In the previous section, we presented a new interpretation of visual prompting: the visual prompting classifier at the later part of the training stage is approximately an adaptive randomized smoothing classifier. In this section, we will derive a new statement on the generalization of the visual prompting classifier during the inference stage: the visual prompting classifier provides a robustness guarantee against perturbation $\delta$ within the radius $R$ for a given sample $x$. In other words, it preserves the most probable class $c$ for the sample $x+\delta$, which a smoothed classifier $\mathcal{G}_{\phi^{*}}(x)$ returns on the input $x$. Let $c_A$ be the most probable class for given $x$, and $p_A$ be the corresponding output probability of the classifier $\mathcal{F}$. The second most probable class is defined similarly as $c_B$ with the corresponding probability to $p_B$. The probabilities of $p_A$ and $p_B$ vary by given input, so it is useful to bound the range of these probabilities and explain the relationship with the smoothed classifier. We denote $\underline{p_A}$ as the lower bound for $p_A$ and $\overline{p_B}$ as the upper bound for $p_B$ as in \cite{cohen2019certified}. Now, assume there exists $c_A \in \mathcal{C}$ and $\underline{p_A},\overline{p_B}$  such that:
\begin{align}
\begin{aligned}
\probP(\mathcal{F}(x+v_{\phi^*}(x))=c_A)&\geq \underline{p_A}\geq \overline{p_B}\geq \\ \nonumber
&\max_{c\neq c_A} \probP(\mathcal{F}(x+v_{\phi^*}(x))=c).
\end{aligned}
\end{align}
\begin{align*}
\text{Then, } \mathcal{G}_{\phi^*}(x+\delta) &=c_A, \; \forall \delta \in \{\delta:||\delta||_2 < R\} \\
\text{where } R &=\frac{1}{2} \sqrt{\lambda_{\min}} (\Phi^{-1}(\underline{p_A})-\Phi^{-1}(\overline{p_B})).  
\end{align*}
Here, $\Phi$ denotes the cumulative distribution function (CDF) of the standardized normal distribution and $\lambda_{\min}$ refers to the minimum eigenvalue of a covariance matrix $\textrm{Cov}(v_{\phi^*}(x))$, and $\sqrt{\lambda_{\min}}$ is bounded from above by $\min\{\sigma_{i}\}_{i=1}^{d}$ where $\sigma_i^2$ is the $i^{th}$-diagonal component of $\textrm{Cov}(v_{\phi^*}(x))$ from the Rayleigh quotient, elaborated in Supplementary. 

\noindent{\textbf{Remark 1) Generalization of Visual Prompting Classifiers. }}From above statement, we observed that visual prompting classifier\footnote{Given a test sample, VP and BlackVIP conduct inference with a single prompt in practice, but one can pursue a more accurate approximation of the behavior of smoothed classifier by storing SGD samples from the few last iterations and averaging the classifier's predictions over them.} $\mathcal{G}_{\phi^{*}}$ is robust around $x+v_{\phi^{*}}(x)$ within the radius $R$, which is impacted by the minimum variance of visual prompt $\min\{\sigma_{i}\}_{i=1}^{d}$ and the difference between $\underline{p_A}$ and $\overline{p_B}$. Importantly, this robustness guarantee is distinctive from that of randomized smoothing classifier $\mathcal{G}$, which ensures the robust prediction around $x$. That is, $\mathcal{G}_{\phi^{*}}$ first move the given sample $x$ with a learned offset direction $v_{\phi^{*}}(x)$ and holds the prediction of base classifier $\mathcal{F}(x+v_{\phi^{*}}(x))$ for the prompt-augmented sample $x+v_{\phi^{*}}(x)$ against a perturbation $\delta$, so that it can fix the incorrect prediction of base classifier $\mathcal{F}(x)$ by a prompt $v_{\phi}(x)$ learned to minimize the classification loss. Then, if we assume the test sample is a perturbed version of a train sample, i.e., $x_{\text{test}}=x_{\text{train}}+\delta$, and both have the same ground truth class label $c$, our $\mathcal{G}_{\phi^{*}}$ ensure generalization of the test sample within $||\delta||_{2}<R$ as it do on the unperturbed train sample, i.e., $\mathcal{G}_{\phi^{*}}(x_{\text{train}})=\mathcal{G}_{\phi^{*}}(x_{\text{test}})=c$. Here, the radius to ensure such generalization is determined by how large the minimum variance of a learned visual prompt is and how large the minimum gap between the top two confidence of the base classifier ($\underline{p_A}-\overline{p_B}$) given input $x$.

\noindent{\textbf{Remark 2) Comparison between VP and BlackVIP. }} According to the above analysis, the following three components are crucial to induce successful generalization in visual prompting classifiers: 1) learned offset $v_{\phi^{*}}(x)$, 2) minimum variance among visual prompt elements $\min\{\sigma_{i}\}_{i=1}^{d}$, and 3) the difference between output probabilities of the most and second-most probable classes from base classifier $\mathcal{F}$. Because the third component is shared for VP and BlackVIP, we focus on the first two components for comparison of VP and BlackVIP in terms of their generalization capability. Upon learning offset $v_{\phi^{*}}(x)$, VP uses a universal prompt $v_{\phi^{*}}(x)=\phi$ regardless of input $x$ while BlackVIP adopts an input-dependent prompt $v_{\phi^{*}}(x)$ which depends on the input $x$ during both the train-time and the test-time. Therefore, our BlackVIP has significantly better capability to modify the incorrect classification of the base model on train samples, $\mathcal{F}(x_{\text{train}})$, and results in generalization for test sample $\mathcal{F}(x_{\text{train}}+\delta)$. Moreover, the radius of robustness guarantee is much more flexible in BlackVIP due to its input-dependency on the variance of visual prompt. We speculate that this kind of conditional coverage of BlackVIP assists flexible generalization depending on the samples' intrinsic variation, and results in better transfer learning performance across diverse environments.

In Figure \ref{fig:thm_illu}, we compare the generalization behavior of RS, VP, and BlackVIP with an illustration of the classifier decision boundary (b), and empirical validation on EuroSAT 16-shot classification task under a white-box true gradient setting (c). Although RS achieves strong robustness to $L_{2}$-noise , which is expected given its theoretical property, RS cannot improve the false predictions of the base classifier. Meanwhile, VP and BlackVIP can adapt the base classifier with a few in-distribution samples, and both successfully improve its in-distribution generalization. BlackVIP shows better $L_{2}$-robustness compared to VP in practice due to its flexible prompt design principle.

\section{Conclusion}
We pioneer \textit{black-box visual prompting} for a realistic and robust adaptation of PTMs. We propose BlackVIP and its efficient alternative, BlackVIP-SE, which reparameterizes the visual prompt with a conditional prompt generation network, Coordinator. By equipping our new optimizer, SPSA-GC, BlackVIPs do not require any access to model architecture or parameters and efficiently adapt the pre-trained vision model to various downstream tasks. Extensive empirical results on 19 datasets show that BlackVIP and BlackVIP-SE consistently achieve performance gains over the baseline on few-shot adaptation in the wilds, with minimal computational cost, memory capacity, and API queries. We further provide theoretical analyses of the generalization of visual prompting methods by providing a novel connection between the certified robustness of randomized smoothing and visual prompting.

\section*{Acknowledgments}
This work was supported by the National Research Foundation of Korea(NRF) grant funded by the Korea government(MSIT)(RS-2024-00457216)
\bibliographystyle{IEEEtran}
\bibliography{refpami}






\appendix
\section{Details of Theoretical Analysis} \label{sec:appendix_thm}
\subsection{Normal Approximation}

The approximation relating parameter updates to sampling from a normal distribution arises from the inherent stochasticity introduced by mini-batch updates in estimating the gradient of the loss function. According to the Central Limit Theorem (CLT), this gradient approaches a normal distribution as the batch size $S$ increases. The application of the CLT relies on the assumption that the stochastic gradient is the summation of independent and uniformly sampled $S$. This leads to the following convergence in distribution for the gradient noise:
\begin{align*}
\nabla \hat{L}_S(\phi)-\nabla L (\phi) \xrightarrow{d} \mathcal{N}(0, \frac{1}{S} C(\phi))
\end{align*}
where $\nabla\hat{L}_S (\phi)$ denotes the stochastic gradient estimate based on a mini-batch of size $S$, $\nabla {L}(\phi)$ represents the true gradient over the entire dataset. Furthermore, we assume the gradient noise, $\nabla \hat{L}_S(\phi)-\nabla L(\phi)$, has a covariance approximated by a constant matrix $\mathbf{C}=\mathbf{B}\mathbf{B}^T$, where $\mathbf{C}$ is factorized into $\mathbf{B}\mathbf{B}^T$. 

\begin{assumption}[Gaussian Gradient Noise with Constant Covariance]
    In the vicinity of a local optimum, assuming a sufficiently small learning rate $\alpha$ such that iterates $\phi_k$ remain confined to a neighborhood, the gradient noise covariance matrix is approximated by a constant, semi-positive definite matrix $C$.
\end{assumption}

The standard discrete-time SGD update rule is given by:
\begin{align*}
    \phi_{k+1} =\phi_k- \alpha \nabla \hat{L}_S(\phi_k)
\end{align*}
where $\phi_k$ denotes the parameter vector at iteration $k$.

\begin{assumption}[Continuous-Time Approximation]
    For a sufficiently small learning rate $\alpha$, the discrete-time SGD is approximated by a continuous-time stochastic differential equation(SDE).
\end{assumption}

\begin{assumption}[Quadratic Loss Approximation]
    Around a local minimum $\mu_{\phi^*}$, the loss function $L(\phi)$ can be approximated by a quadratic function form:
    \begin{align*}
        L(\phi)\approx L(\mu_{\phi^*})+\frac{1}{2}(\phi-\mu_{\phi^*})^T \mathbf{A} (\phi-\mu_{\phi^*}).
    \end{align*}

    Without loss of generality, the minimum is located at $\phi=0$, which simplifes the equation to
    \begin{align*}
        L(\phi)\approx \frac{1}{2}\phi ^T \mathbf{A} \phi.
    \end{align*}
\end{assumption}
Here, we also assume positive definiteness on $\mathbf{A}$. These assumptions come from \cite{stephan2017stochastic} and enables the continuous-time SDE of $\phi$ to reduce to a multivariate Ornstein-Uhlenbeck process, which admits a Gaussian stationary distribution, where: 
\begin{align*}
q(\phi)\propto \exp\{-\frac{1}{2}\phi^T\Sigma_{\phi}^{-1}\phi\}.
\end{align*}

$q(\phi)$ denotes a stationary distribution and the covariance $\Sigma_{\phi}$ satisfies $\Sigma_{\phi} \mathbf{A}+\mathbf{A}\Sigma_{\phi} = \frac{\alpha}{S}\mathbf{B}\mathbf{B}^T$, with $\mathbf{A}$ being the hessian matrix at the optimal point. 

\noindent
\textbf{Normal Approximation of SPSA/SPSA-GC.} To remove the white-box dependency, we rewrite SPSA (and SPSA-GC) as a stochastic approximation recursion driven by a martingale-difference noise, and invoke a martingale functional CLT (MFCLT) to justify the Brownian-motion approximation.

\begin{assumption}[SPSA late-stage regularity and stability]\label{ass:spsa_reg}
In a neighborhood of a local minimizer $\mu_{\phi^*}$:
(i) $L$ is three-times continuously differentiable and its third derivatives are bounded.
(ii) We consider a late-stage regime where $\alpha$ is constant and small,
and $c_k\equiv c$ (or slowly varying so that $O(c_k^2)$ is small).
(iii) Stable : the induced linear drift is stable, i.e., $\rho(I-\alpha \mathbf{A})<1$ which means the spectral radius of $I-\alpha \mathbf{A}$ is less than 1. 

\end{assumption}

SPSA update has the following form. 
\begin{align*}
\hat g_k(\phi_k)
&=\frac{L(\phi_k+c_k\Delta_k)-L(\phi_k-c_k\Delta_k)}{2c_k}\,\Delta_k^{-1}, \\
\phi_{k+1}&=\phi_k-\alpha\,\hat g_k(\phi_k),
\end{align*}
where $\Delta_k$ is a vector with i.i.d. Rademacher components. Furthermore, SPSA-GC update has the following form.
\begin{align*}
m_{k+1}=\beta m_k-\alpha\,\hat g_k(\phi_k+\beta m_k),
\quad
\phi_{k+1}=\phi_k+m_{k+1},
\end{align*}
matching Algorithm~1 in the main paper.

Let $\mathcal F_k\triangleq\sigma(\phi_0,\Delta_1,\ldots,\Delta_k)$ (so $\phi_k$ is $\mathcal F_{k-1}$-measurable).
Define
\begin{align*}
b_k &= \mathbb E[\hat g_k(\phi_k)\mid \mathcal F_{k-1}]-\nabla L(\phi_k),\\
\xi_k &= \hat g_k(\phi_k)-\mathbb E[\hat g_k(\phi_k)\mid \mathcal F_{k-1}],
\end{align*}
so $\mathbb E[\xi_k\mid\mathcal F_{k-1}]=0$ and $\{\xi_k,\mathcal F_k\}$ is an martingale difference sequence (MDS). We define the shorthand
\begin{align*}
    T(\phi,\Delta)&= \Delta^\top \nabla^2 L(\phi)\,\Delta,
\end{align*}
Using two-sided Taylor expansions,
{\small
\begin{align*}
L(\phi_k\!+\!c_k\Delta_k)
&=L(\phi_k)+c_k\Delta_k^\top\nabla L(\phi_k)+\frac{c_k^2}{2}T(\phi_k,\Delta_k)
+O(c_k^3) \\
L(\phi_k\!-\!c_k\Delta_k)
&=L(\phi_k)-c_k\Delta_k^\top\nabla L(\phi_k) +\frac{c_k^2}{2}T(\phi_k,\Delta_k)
+O(c_k^3),
\end{align*}}
and substituting into $\hat g_k(\phi_k)$ yields
\begin{align*}
\hat g_k(\phi_k)
&=\Delta_k\Delta_k^\top \nabla L(\phi_k)
+O(c_k^2)\\
&=\nabla L(\phi_k)+(\Delta_k\Delta_k^\top-I)\nabla L(\phi_k)+O(c_k^2).
\end{align*}
Since $\mathbb E[\Delta_k\Delta_k^\top]=I$,
\begin{align*}
\mathbb E[\hat g_k(\phi_k)\mid\mathcal F_{k-1}]
&=\nabla L(\phi_k)+O(c_k^2),
\quad
b_k=O(c_k^2),
\\
\mathbb E[\xi_k\mid\mathcal F_{k-1}]&=0.
\end{align*}

Because $\Delta_k$ has bounded moments (e.g.\ $\pm1$ entries) and we stay in a local neighborhood (compact set), $\xi_k$ has finite conditional second moment, i.e., $\mathbb{E}[\Vert \xi_k\Vert ^2 | \mathcal{F}_{k-1}]<\infty$; hence $\Sigma_\xi=\mathbb E[\xi_k\xi_k^\top]$ exists. If $c_k= c$ in the late stage, then $b_k\approx b$ with $\|b\|=O(c^2)$,
and the mean fixed point is $\bar\phi\approx \mu_{\phi^*} -A^{-1}b$.
Let $e_k=\phi_k-\bar\phi$ and $\tilde e_k=e_k-\mathbb E[e_k]$. Then
\[
\tilde e_{k+1}\approx (I-\alpha \mathbf{A})\tilde e_k-\alpha\,\xi_k.
\]
Moreover, cross terms vanish since $\tilde e_k$ is $\mathcal F_{k-1}$-measurable and
$\mathbb E[\xi_k\mid\mathcal F_{k-1}]=0$:
\[
\mathbb E[\tilde e_k\xi_k^\top]
=\mathbb E\!\big[\tilde e_k\,\mathbb E[\xi_k^\top\mid\mathcal F_{k-1}]\big]=0.
\]
Hence the stationary covariance solves the discrete Lyapunov equation
(which has a unique positive semi-definite solution when $\rho(I-\alpha \mathbf{A})<1$):
\begin{align*}
\Sigma_\phi = \mathbf{M}\Sigma_\phi \mathbf{M}^\top+\alpha^2\Sigma_\xi,
\quad
\mathbf{M} = I-\alpha \mathbf{A}.
\end{align*}
Under the stable linearized model, this implies the Markov chain admits a unique stationary law
(with finite second moment), centered at $\bar\phi$.

\begin{assumption}[MFCLT conditions for $\{\xi_k,\mathcal F_k\}$]\label{ass:mfclt}
There exists a positive semi-definite matrix $\Sigma_\xi$ such that for each fixed $t\ge0$,
(i) the predictable quadratic variation stabilizes:
\[
\alpha\sum_{i=1}^{\lfloor t/\alpha\rfloor}
\mathbb E[\xi_i\xi_i^\top\mid\mathcal F_{i-1}]
\ \xrightarrow{\ \mathbb P\ }\ t\,\Sigma_\xi,
\]
(ii) and the Lindeberg condition holds:
\[
\alpha\sum_{i=1}^{\lfloor t/\alpha\rfloor}
\mathbb E\!\left[\|\xi_i\|^2\mathbf 1\{\|\xi_i\|>\varepsilon/\sqrt{\alpha}\}\mid\mathcal F_{i-1}\right]
\ \xrightarrow{\ \mathbb P\ }\ 0
\quad(\forall\varepsilon>0).
\]
\end{assumption}
Then the scaled partial-sum process converges to Brownian motion:
\begin{align*}
\sqrt{\alpha}\sum_{i=1}^{\lfloor t/\alpha\rfloor}\xi_i
\ \Rightarrow\ \Sigma_\xi^{1/2}W(t),
\end{align*}
where $W(t)$ is a standard Brownian motion.

Consequently, the scaled interpolation
$u^{(\alpha)}(t)\triangleq \tilde e_{\lfloor t/\alpha\rfloor}/\sqrt{\alpha}$
converges (as $\alpha\to0$) to the OU SDE
\begin{align*}
du(t)=-\mathbf{A}\,u(t)\,dt+\Sigma_\xi^{1/2}\,dW(t).
\end{align*}
The OU process admits a Gaussian stationary distribution
$u\sim\mathcal N(0,\Sigma_u)$ with $\mathbf{A}\Sigma_u+\Sigma_u\mathbf{A}^\top=\Sigma_\xi$,
hence in the late-stage small-$\alpha$ regime,
$\phi_k\approx \mathcal N(\bar\phi,\alpha\Sigma_u)$.

SPSA-GC also shows similar results. Let $s_k \triangleq [\tilde e_k^\top,\, m_k^\top]^\top$.
Linearizing $\nabla L(\phi_k+\beta m_k) \approx \mathbf{A}(\tilde e_k+\beta m_k)$ yields the stable linear recursion
\begin{align*}
s_{k+1} & \approx G s_k - \alpha
\begin{bmatrix}
\xi^*_k\\ \xi^*_k
\end{bmatrix},
\quad
G=
\begin{bmatrix}
I-\alpha \mathbf{A} & \beta I-\alpha\beta \mathbf{A}\\
-\alpha \mathbf{A} & \beta I-\alpha\beta \mathbf{A}
\end{bmatrix} ,\\
\xi^*_k&=\hat{g}_k(\phi_k+\beta m_k)-\mathbb{E}[\hat{g}_k(\phi_k+\beta m_k)| \mathcal{F}_{k-1}]. 
\end{align*}
If $\rho(G)<1$ and $\mathbb{E}\|\xi^*_k\|^2<\infty$, the linearized chain admits a stationary regime with finite second moments,
and the stationary covariance $\Sigma_s=\mathbb{E}[s_k s_k^\top]$ satisfies the discrete-time block Lyapunov equation
\begin{align*}
\Sigma_s = G \Sigma_s G^\top + \alpha^2 \Sigma_\eta,
\quad
\Sigma_\eta =
\mathbb{E}\!\left[
\begin{bmatrix}\xi^*_k\\ \xi^*_k\end{bmatrix}
\begin{bmatrix}\xi^*_k\\ \xi^*_k\end{bmatrix}^{\!\top}
\right]
=
\begin{bmatrix}
\Sigma_{\xi^*} & \Sigma_{\xi^*}\\
\Sigma_{\xi^*} & \Sigma_{\xi^*}
\end{bmatrix}.
\end{align*}
Moreover, in the late-stage small-step regime, $m_k$ is a fast stable component and can be eliminated to first order, yielding
the effective recursion
\begin{align*}
\tilde e_{k+1} \approx (I-\bar\alpha \mathbf{A})\tilde e_k - \bar\alpha\,\xi^*_k,
\quad
\bar\alpha =\frac{\alpha}{1-\beta}.
\end{align*}
Therefore, by Assumption~\ref{ass:mfclt} (MFCLT) for $\{\xi^*_k\}$, the scaled error process admits an
Ornstein--Uhlenbeck (OU) diffusion approximation, which implies an (approximately) Gaussian stationary law for $\tilde e_k$.
In particular, letting $\Sigma_{u^*}$ solve
\begin{align*}
\mathbf{A}\Sigma_{u^*}+\Sigma_{u^*} \mathbf{A}^\top = \Sigma_{\xi^*},
\end{align*}
we obtain the late-stage Gaussian approximation
\begin{align*}
\phi_k \approx \mathcal{N}\!\Big(\bar\phi,\; \bar\alpha\,\Sigma_{u^*}\Big)
= \mathcal{N}\!\Big(\bar\phi,\; \tfrac{\alpha}{1-\beta}\,\Sigma_{u^*}\Big).
\end{align*}

\subsection{Randomized Smoothing and Certified Robustness}
A smoothed classifier exhibits robust properties within a radius $R$. Specifically, when an input is perturbed within the ball $B(x,R)=\{y:||x-y||<R\}$, the classification consistently predicts the most probable class. Building upon the foundational work by \cite{cohen2019certified} which introduced a certified radius for a smoothed classifier using an isotropic Gaussian covariance, and further extensions by \cite{hong2022certified} to an anisotropic Gaussian model with a diagonal covariance matrix, our analysis advances these models by relaxing the diagonal constraint on the covariance matrix, thus proposing a more generalized form and redefining the radius based on the derived proof. 

To set the stage for our proof, consider the following settings:
\begin{align*}
    & X=x+\phi^* \sim \mathcal{N}_d(x+\mu_{\phi^*},\Sigma_{\phi^*}) \\
    & X^*=x+\phi^*+\delta \sim \mathcal{N}_d(x+\mu_{\phi^*}+\delta,\Sigma_{\phi^*}).
\end{align*}
We examine the ratio of the probability density functions of $X^*$ and $X$, which manifests in an exponential form with the precision matrix $K=\Sigma_{\phi^*}^{-1}$. The ratio $\frac{p_{X^*}(z)}{p_{X}(z)}$ is given by:
\begin{align*}
& \frac{\exp\{-\frac{1}{2}(z-(x+\mu_{\phi^*}+\delta))^TK(z-(x+\mu_{\phi^*}+\delta))\}}{\exp\{-\frac{1}{2}(z-(x+\mu_{\phi^*}))^TK(z-(x+\mu_{\phi^*}))\}} \\
&= \exp\{\frac{1}{2}(2\delta^TKz-2\delta^TK(x+\mu_{\phi^*})-\delta^TK\delta)\} \\
&= \exp\{\delta^TKz-\delta^TK(x+\mu_{\phi^*})-\frac{1}{2}\delta^TK\delta\} \\
&= \exp\{\delta^TKz-\tau\}
\end{align*}
where $\tau=\delta^TK(x+\mu_{\phi^*})+\dfrac{1}{2}\delta^TK\delta$. This leads us to the following inequality:
\begin{align*}
    \exp\{\delta^TKz-\tau\}\leq w \Longleftrightarrow
    \delta^TKz \leq \log{w}+\tau=\eta.
\end{align*}
Leveraging the Neyman-Pearson theorem in \cite{cohen2019certified}, we establish that if the following condition holds for any binary classifier $\mathcal{F}_{\textrm{bin}}:\mathbb{R}^d \rightarrow \{0,1\}$:
\begin{align*}
& \probP(\mathcal{F}_{\textrm{bin}}(X)=1)\geq \probP(X\in \Omega) \\
\Omega=\{z\in& \mathbb{R}^d: \frac{p_{X^*}(z)}{p_X(z)}\leq w \} \textrm{ for some } w>0
\end{align*}
then it must hold that $\probP(\mathcal{F}_{\textrm{bin}}(X^*)=1)\geq \probP(X^*\in \Omega)$. This statement is also satisfied when the direction of the inequality is reversed:
\begin{align*}
& \probP(\mathcal{F}_{\textrm{bin}}(X)=1)\leq \probP(X\in \Omega) \\
\Omega=\{z\in& \mathbb{R}^d: \frac{p_{X^*}(z)}{p_X(z)}\geq w \} \textrm{ for some } w>0
\end{align*}
indicating that $\probP(\mathcal{F}_{\textrm{bin}}(X^*)=1)\leq \probP(X^*\in \Omega)$ is also true. This principle underpins our main claim. Suppose there exists a class $c_A$ within the set $\mathcal{C}$ with probability $p_A$ bounded by $\underline{p_A}$ and $\overline{p_B}$ such that:
\begin{align*}
\probP(\mathcal{F}(x+\phi^*)=c_A)&\geq \underline{p_A}\geq \overline{p_B}\geq \\
&\max_{c\neq c_A} \probP(\mathcal{F}(x+\phi^*)=c).    
\end{align*}
Then, it necessarily follows that:
\begin{align*}
  \mathcal{G}_{\phi^*}(x+\delta) &=c_A, \; \forall \delta \in \{\delta:||\delta||_2 < R\} \\
\text{where } R &=\frac{1}{2} \sqrt{\lambda_{\min}} (\Phi^{-1}(\underline{p_A})-\Phi^{-1}(\overline{p_B})).  
\end{align*}
The result means that our smoothed classifier $\mathcal{G}_{\phi^*}$ retains the most probable class of the original classifier even in the presence of input perturbation. Formally, it implies that $\probP(\mathcal{F}(X^*)=c_A)$ is at least as large as $\probP(\mathcal{F}(X^*)=c)$ for any $c\in\mathcal{C}-\{c_A\}$. Without loss of generality, we fix $c$ as $c_B$ which is the second most probable class, then we need to show that $\probP(\mathcal{F}(X^*)=c_A)\geq \probP(\mathcal{F}(X^*)=c_B)$. The bounds for probabilities $\underline{p_A}$ and $\overline{p_B}$ can be expressed as $\probP(X\in \Omega_A)$ and $\probP(X\in \Omega_B)$ when the regions $\Omega_A$ and $\Omega_B$ are given by:
\begin{align*}
    & \Omega_A=\{z:\delta^TK(z-(x+\mu_{\phi^*}))^T \leq \kappa\Phi^{-1}
    (\underline{p_A})\} \\
    & \Omega_B=\{z:\delta^TK(z-(x+\mu_{\phi^*}))^T \geq \kappa\Phi^{-1}
    (1-\overline{p_B})\}.  \\
\end{align*}
In defining these regions, we denote $\kappa$ as the norm $=||\delta^T\Sigma_{\phi^*}^{-1/2}||_2.$ To further clarify the discussion, we define several useful random variables transformed from $X$:
\begin{align*}
     & Y=X-(x+\mu_{\phi^*}) \sim \mathcal{N}_d(0,\Sigma_{\phi^*}) \\
     & Y^*=X-(x+\mu_{\phi^*}+\delta) \sim \mathcal{N}_d(\delta,\Sigma_{\phi^*}) \\
     & \Xi=\Sigma_{\phi^*}^{-1/2}Y \sim \mathcal{N}_d(0,\mathbf{I}) \\
     & Z \sim \mathcal{N}(0,1).
 \end{align*}
 We show $\probP(X\in \Omega_A)=\underline{p_A}$ as follows:
\begin{align*}
  \probP(X\in \Omega_A)
  &=\probP(\delta^T K (X-x-\mu_{\phi^*}) \leq \kappa\Phi^{-1}(\underline{p_A}))\\
  &=\probP(\delta^T K Y \leq \kappa\Phi^{-1}(\underline{p_A})) \\
  &=\probP(\delta^T K \Sigma_{\phi^*}^{1/2} \Xi \leq \kappa\Phi^{-1}(\underline{p_A})) \\
  &=\probP(\delta^T \Sigma_{\phi^*}^{-1/2} \Xi \leq \kappa\Phi^{-1}(\underline{p_A})) \\
  &=\probP(||\delta^T \Sigma_{\phi^*}^{-1/2}||_2 Z\leq \kappa\Phi^{-1}(\underline{p_A})) \\
  &=\probP(Z\leq \Phi^{-1}(\underline{p_A})) \\
  &=\underline{p_A}.
\end{align*}
Here, $\delta^T\Sigma_{\phi^*}^{-1/2}$ is a row vector in $\mathbb{R}^{1\times d}$ and a random vector $\Xi$ in $\mathbb{R}^{d\times 1}$ follows a multivariate standard normal distribution. The inner product of theses vectors results in a weighted sum of univariate standard normal distributions, which reduces to a univariate normal distribution scaled by the constant $\kappa$. For $\Omega_B$, the probability is calculated as:
\begin{align*}
  \probP(X\in \Omega_B)
  &=\probP(\delta^T K (X-x-\mu_{\phi^*}) \geq \kappa\Phi^{-1}(1-\overline{p_B}))\\
  &=\probP(\delta^T K Y \geq \kappa\Phi^{-1}(1-\overline{p_B})) \\
  &=\probP(\delta^T K \Sigma_{\phi^*}^{1/2} \Xi \geq \kappa\Phi^{-1}(1-\overline{p_B})) \\
  &=\probP(\delta^T \Sigma_{\phi^*}^{-1/2} \Xi \geq \kappa\Phi^{-1}(1-\overline{p_B})) \\
  &=\probP(||\delta^T \Sigma_{\phi^*}^{-1/2}||_2Z \geq \kappa \Phi^{-1}(1-\overline{p_B}))\\
  &=\probP(Z\geq \Phi^{-1}(1-\overline{p_B})) \\
  &=\overline{p_B}.
\end{align*}
We can modify our initial assumption by replacing $\underline{p_A}$ to $\probP(X\in \Omega_A)$ and $\overline{p_B}$ to $\probP(X\in \Omega_B)$. 
\begin{align*}
    \probP(\mathcal{F}(X)=c_A)\geq \probP(X\in \Omega_A) \\
    \probP(\mathcal{F}(X)=c_B)\leq \probP(X\in \Omega_B).
\end{align*}
We utilize Neyman-Pearson theorem by defining $\mathcal{F}_{\textrm{bin}}(z)=I[\mathcal{F}(z)=c_A]$ where $I$ is an indicator function, then the result are ensured
\begin{align*}
    \probP(\mathcal{F}(X^*)=c_A)\geq \probP(X^*\in \Omega_A) \\
    \probP(\mathcal{F}(X^*)=c_B)\leq \probP(X^* \in \Omega_B).
\end{align*}
If $\probP(X^*\in \Omega_A)\geq \probP(X^* \in \Omega_B)$ is satisfied, $\probP(\mathcal{F}(X^*)=c_A)\geq \probP(\mathcal{F}(X^*)=c_B)$ is held with a certain radius, which is the destination of our proof.
\begin{align*}
    \probP(\mathcal{F}(x+\phi+\delta)=c_A)\geq \max_{c\neq c_A} \probP(\mathcal{F}(x+\phi+\delta)=c).
\end{align*}
To show the inequality, we plug in the space $\Omega_A$ we defined into $\probP(X^*\in \Omega_A)$
\begin{align*}
    & \probP(X^* \in \Omega_A) \\
    &= \probP(\delta^T K(X^*-x-\mu_{\phi^*})\leq \kappa\Phi^{-1}(\underline{p_A})) \\
    &= \probP(\delta^T K Y^* \leq \kappa\Phi^{-1}(\underline{p_A})) \\
    &= \probP(\delta^TK Y + \delta^TK\delta \leq \kappa\Phi^{-1}(\underline{p_A})) \\ 
    &= \probP(\delta^T\Sigma_{\phi^*}^{-1/2}\Xi \leq \kappa\Phi^{-1}(\underline{p_A}) - \delta^TK\delta) \\
    &= \probP(\kappa Z \leq \kappa (\Phi^{-1}(\underline{p_A})-\frac{1}{\kappa}\delta^TK\delta)) \\
    & = \probP(Z\leq \Phi^{-1}(\underline{p_A})-\frac{1}{\kappa}\delta^TK\delta) \\
    & = \Phi(\Phi^{-1}(\underline{p_A})-\frac{1}{\kappa}\delta^TK\delta).
\end{align*}
We follow a similar procedure for the space $\Omega_B$
\begin{align*}
    & \probP(X^*\in \Omega_B) \\
    &= \probP(\delta^T K(X^*-x-\mu_{\phi^*})\geq \kappa\Phi^{-1}(1-\overline{p_B})) \\
    &= \probP(\delta^T K Y^* \geq \kappa\Phi^{-1}(1-\overline{p_B})) \\
    &= \probP(\delta^TK Y + \delta^TK\delta \geq \kappa\Phi^{-1}(1-\overline{p_B})) \\ 
    &= \probP(\delta^T\Sigma_{\phi^*}^{-1/2}\Xi \geq \kappa\Phi^{-1}(1-\overline{p_B}) - \delta^TK\delta) \\
    &= \probP(\kappa Z \geq \kappa (\Phi^{-1}(1-\overline{p_B})-\frac{1}{\kappa}\delta^TK\delta)) \\
    &= \probP(Z \geq \Phi^{-1}(1-\overline{p_B})-\frac{1}{\kappa}\delta^TK\delta) \\
    & = \probP(Z\leq \Phi^{-1}(\overline{p_B})+\frac{1}{\kappa}\delta^TK\delta) \\
    & = \Phi(\Phi^{-1}(\overline{p_B})+\frac{1}{\kappa}\delta^TK\delta).
\end{align*}
By comparing these two probabilities, we identify the condition under which $\probP(X^*\in \Omega_A)\geq \probP(X^*\in \Omega_B)$ holds as follows:
\begin{align*}
& \Phi(\Phi^{-1}(\underline{p_A})-\frac{1}{\kappa}\delta^TK\delta) \geq \Phi(\Phi^{-1}(\overline{p_B})+\frac{1}{\kappa}\delta^TK\delta)  \\
& \Longleftrightarrow  \quad \Phi^{-1}(\underline{p_A})-\frac{1}{\kappa}\delta^TK\delta \geq \Phi^{-1}(\overline{p_B})+\frac{1}{\kappa}\delta^TK\delta \\
&\Longleftrightarrow  \quad    \frac{\delta^TK\delta}{\kappa} \leq \frac{1}{2}(\Phi^{-1}(\underline{p_A})-\Phi^{-1}(\overline{p_B})).
\end{align*}
$\kappa$ is previously defined as $||\delta^T\Sigma^{-1/2}_{\phi^*}||_2$, which is $\sqrt{\delta^T\Sigma^{-1}_{\phi^*}\delta}$. Since $\Sigma_{\phi^*}$ is a Hermitian matrix, $K=\Sigma^{-1}_{\phi^*}$ is also a Hermitian matrix, satisfying the condition for Rayleigh quotient in \cite{horn2012matrix}. The upper bound of $\frac{\delta^TK\delta}{\kappa}$ is $\frac{||\delta||}{\sqrt{\lambda_{\min}}}$, as determined by Rayleigh quotient: 
\begin{align*}
    \frac{\delta^TK\delta}{\kappa}=\frac{\delta^T\Sigma^{-1}_{\phi^*}\delta}{\sqrt{\delta^T\Sigma^{-1}_{\phi^*}\delta}}=\sqrt{\delta^T\Sigma^{-1}_{\phi^*}\delta}\leq \sqrt{\frac{||\delta||^2}{\lambda_{\min}}}=\frac{||\delta||}{\sqrt{\lambda_{\min}}}.
\end{align*}
Here, $\lambda_{\min}$ stands for the minimum eigenvalue of $\Sigma_{\phi^*}$. The right side is derived by the equality $\lambda_{\max}(\Sigma_{\phi^*}^{-1})=\frac{1}{\lambda_{\min}(\Sigma_{\phi^*})}$, where $\lambda_{\max}(\Sigma_{\phi^*}^{-1})$ denotes the maximum eigenvalue of the matrix $\Sigma^{-1}_{\phi^*}$, and $\lambda_{\min}(\Sigma_{\phi^*})$ is the minimum eigenvalue of $\Sigma_{\phi^*}$ denoted as $\lambda_{\min}$. When the upper bound is less than or equal to $\frac{\kappa}{2}\Psi$  such that $\Psi=\Phi^{-1}(\underline{p_A})-\Phi^{-1}(\overline{p_B})$,  $\mathcal{G}_{\phi^*}(x+\delta)=c_A$ is supported.
\begin{align*}
    \frac{||\delta||}{\sqrt{\lambda_{\min}}} \leq \frac{1}{2}\Psi
\Longleftrightarrow ||\delta||\leq \frac{1}{2}\sqrt{\lambda_{\min}}\Psi=R.
\end{align*}
In other words, our classifier with Gaussian visual prompt ensures the robustness against the perturbation $\delta$ by $R$. Also, for any vector $x$ in $\mathbb{R}^d$, the upper bound of $\lambda_{\min}$ is introduced by Rayleigh quotient.
\begin{align*}
    \lambda_{\min} & \leq \min_{x\neq 0}{\frac{x^T\Sigma_{\phi^*}x}{x^Tx}} \\ 
    &=\min_{\sum^d_{i=1}x_i^2=1}{(x_1^2\sigma_1^2+\cdots+x_d^2\sigma_d^2)} \\
    &=\min\{\sigma_i^2\}^d_{i=1}.
\end{align*}
This induces the upper bound of $R$, which is $\overline{R}=\frac{1}{2}\min\{\sigma_i\}^d_{i=1}\Psi$. 

\section{Experimental Setting}
\label{app:b_exp_setup}
\subsection{Datasets}
\label{app:b_dataset}
\begin{figure}[!htp]
     \centering
     \subfloat[]{\includegraphics[width=0.5\textwidth]{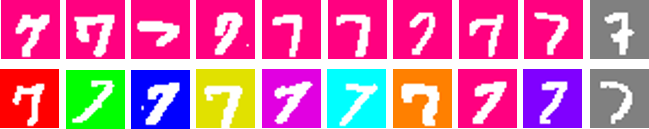}%
     \label{fig:a_bmnist}}
    \hfill
    \subfloat[]{\includegraphics[width=0.5\textwidth]{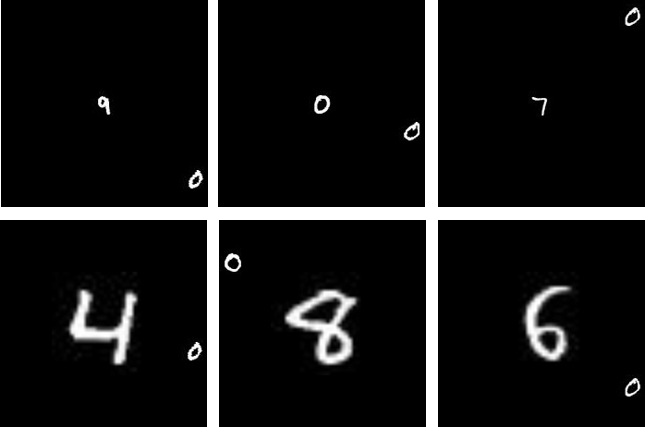}%
    \label{fig:a_lmnist}}
    \caption{Examples of two synthetic datasets. (a) Examples of $y=7$ subset in Biased-MNIST \cite{bahng2020learning} with $\rho=0.9$. (Top) The train set is constructed with the spurious correlation between the background color and digit class (e.g., $y=7$ occurs $90\%$ with a pink background and $10\%$ with other random colors in this case). (Bottom) The test set is constructed with a reversed correlation to that of the train set (e.g., $y=7$ occurs $10\%$ with a pink background and $90\%$ with other random colors in this case); and (b) Examples of the Loc-MNIST dataset. The real digit from MNIST is located in the outer area, while the fake digit from another random MNIST image is placed in the center of the image. (Top) The case where the size ratio of the real digit to the fake digit is 1:1, and (Bottom) 1:4.}
    \label{fig:a_synthetic_examples}
\end{figure}

\paragraph{Synthetic Datasets}
Our BlackVIP generates the input-dependent image-size visual prompt which covers the whole image region, so we expect that this flexible prompt design can improve some kind of robustness as well as general recognition capability: (1) To evaluate the robustness on distribution shift (i.e., domain generalization), we consider the Biased-MNIST \cite{bahng2020learning} dataset. (2) To evaluate the robustness on adversarial noise and location-agnostic recognition capacity, we create a variant of the MNIST dataset called Loc-MNIST. Examples of these two datasets are provided in Figure \ref{fig:a_synthetic_examples}.

\textbf{Biased MNIST} is a modified version of MNIST \cite{lecun1998gradient} where the biases reside in the background colors of the images of each digit. At train time, each digit has a unique preassigned background color that strongly correlates with the label. The degree of correlation is determined by the value $\rho \in [0, 1]$, such that $(100 \times \rho) \%$ of the images that belong to the same digit have the preassigned color of that digit as their background color, and the rest are uniformly assigned to have any of the other colors as their background color. At test time, we reverse the ratio so that  $(100 \times (1 - \rho)) \%$ of the images now have the preassigned color as their background color and vice versa to evaluate the model’s dependency on superficial features such as the color of the background that a digit is located on. We prepare the following two environments 1) easy: $\rho=0.8$ and 2) hard: $\rho=0.9$.

On the given black blank image with $224\times224$ resolution, i.e., zero array, \textbf{Loc-MNIST} puts an original target digit image from MNIST that has $28\times28$ resolution on the edge-side (e.g., 0$\sim$27 or 196$\sim$223 for one of the vertical or horizontal sides and 0$\sim$223 for the other side) and puts a random fake digit (also from the MNIST dataset) on the center. The location of the target digit in the edge and the class of the fake digit are chosen randomly with uniform probability. A synthetic image is created one by one for each original MNIST image. We prepare the following two environments 1) easy: the scale of the target and the fake digit is the same, i.e., 1:1, and 2) hard: the fake digit is four times larger than the original digit, i.e., 1:4.

For consistency, we perform the experiments on these two datasets with a few-shot evaluation protocol. To construct a train set, we randomly sample a subset (K-shot) of the created images for each class and use the whole test set.

\paragraph{Datasets}
To extensively evaluate the effectiveness of our proposed method and baseline approaches, we measure performance across the following 14 datasets that are widely used for transfer learning benchmark: Caltech101 \cite{1384978}, OxfordPets \cite{6248092}, StanfordCars \cite{6755945}, Flowers102 \cite{4756141}, Food101 \cite{10.1007/978-3-319-10599-4_29}, FGVCAircraft \cite{maji2013fine}, SUN397 \cite{5539970}, DTD \cite{6909856}, SVHN \cite{37648}, EuroSAT \cite{helber2019eurosat}, Resisc45 \cite{7891544}, CLEVR \cite{johnson2017clevr}, UCF101 \cite{soomro2012ucf101}, and ImageNet (IN) \cite{5206848}. Note that these 14 datasets cover diverse visual domains, and they require understanding various visual semantics like scenes, actions, fine-grained categories, textures, satellite imagery, digits, the number of objects, and the recognition of generic objects.  Moreover, to evaluate the robustness of prompt-based adaptation of PTM in the wild, we additionally consider three benchmark datasets of distribution shifts from WILDS \cite{koh2021wilds}: Camelyon17 \cite{bandi2018detection}, FMoW \cite{christie2018functional}, and iWildCam \cite{beery2021iwildcam}.

Following the protocol in \cite{zhou2022learning, zhou2022conditional}, we conduct a few-shot evaluation for all datasets: 16-shot for the train set, 4-shot for the validation set, and the whole test set. We use the few-shot split by \cite{zhou2022learning} for each dataset, those are also used in \cite{zhou2022learning}, while for Resisc45, CLEVR, and WILDS datasets, we randomly select the 16-shot and 4-shot samples for the training and validation datasets, respectively.

\subsection{Backbone Model}
\label{app:b_backbone}
In this work, we aim at the robust adaptation of pre-trained models on diverse downstream tasks. For these pre-trained models, all experiments in this paper are done with the off-the-shelf vision-language model CLIP \cite{radford2021learning}, and we adopt the ViT-B/16 for image encoder backbone architecture by default. During the adaptation (training) phase, the entire components of the pre-trained model are frozen without any architectural modification, and we only manage and optimize the learnable module Coordinator from the outside of the pre-trained model.

While input space visual prompting allows it to be applied to not only VLM, but also any other vision models like CNNs and ViTs, it requires the user to define the output space mapping, which maps the output prediction category set of a pre-trained task to a new downstream category set \cite{elsayed2018adversarial, tsai2020transfer, bahng2022visual}. This is another non-trivial problem. Therefore, we limit our focus to only the VLM that can dynamically build the task-specific head from a manual text template \cite{radford2021learning, jia2021scaling} so that free from defining output space mapping.

\subsection{Baseline Methods}
\label{app:b_baseline}
\paragraph{CLIP Zero-Shot (ZS)} 
CLIP \cite{radford2021learning} is one of the most popular vision-language zero-shot models that is widely exploited for classification, detection, segmentation, and other vision or vision-language tasks. Based on its well-aligned vision-language joint embedding space, the zero-shot classification can be performed with a manual text prompt (also called a template) of each pre-defined class category. In this paper, we are mainly aiming to improve CLIP's strong zero-shot performance in the few-shot adaptation setting. 

\paragraph{BAR} Black-Box Adversarial Reprogramming (BAR) \cite{tsai2020transfer} was proposed for efficient transfer learning of a pre-trained model to the medical image domain. Different from the previous works on Adversarial Reprogramming (AR), BAR exploits the perturbation-vulnerability of neural networks for \textit{adaptation} purposes rather than attack. By optimizing the frame-shaped learnable program, which embeds a downstream target image inside of that, BAR steers the ImageNet pre-trained model to classify the specialized medical images. Moreover, BAR adopts the zeroth-order optimizer (ZOO), Randomized Gradient-Free (RGF) \cite{10.1007/s10208-015-9296-2} minimization algorithm for black-box transfer learning to broaden its applications.

When the resolution of the downstream input image is over that of the pre-training phase, Tsai et al. \cite{tsai2020transfer} set the embedded target image size for $64\times64$ resolution in the $299\times299$-size learnable program by default. However, we observe that such a heavy-pad thin-image design of the prompt degrade the performance significantly, so we tune the resolution of the embedded image and set $194\times194$.

\paragraph{VP}
Similarly, Visual Prompting (VP) aims at adapting a pre-trained model to downstream tasks via learning input space visual prompts. Among some candidates for prompt designs, Bahng et al. \cite{bahng2022visual} adopt the padding-style prompt so that realized prompts look like the frame-shape program of ARs. VP learns a universal visual prompt per each downstream task, and it just adds to all of the images in a task. Unlike the AR methods or our BlackVIP, the range of prompted images is unbounded. Following \cite{bahng2022visual}, we use the padding-style prompt, which is 30-pixel sized for each side by default.

While VP optimizes the parameters in the input space, it relies on a first-order optimization algorithm that uses the true gradient of the entire model parameters, and we establish the performance of VP as a kind of upper bound for other input space black-box optimization approaches, including BlackVIP and BlackVIP-SE. Additionally, by replacing the first-order algorithm with zeroth-order counterparts, we build two new baselines \textbf{VP w/ SPSA} and \textbf{VP w/ SPSA-GC} on our extensive experiments. These two methods confirm the effectiveness of our new components \textit{Coordinator} and SPSA-GC.

Furthermore, we consider some representative prompt learning methods as baselines on language modality (CoOp \cite{zhou2022coop} and CoCoOp \cite{zhou2022conditional}), vision modality (VP \cite{bahng2022visual} and VPT \cite{jia2022visual}), and both modality (MaPLe \cite{khattak2023maple}) for an evaluation of white-box transfer learning experiments to validate the effectiveness of the design of our visual prompt, solely. All of these methods adapt parameters inside the model, e.g., tokens in the embedding layer or intermediate layers of text or visual encoder of the target vision-language model.

\paragraph{Discussion}
Although BAR, VP, and BlackVIP share the generic goal: efficient transfer learning of pre-trained models via input-space optimization, there are several significant differences. (1) We propose a novel prompt design that is automatically formed in an input-dependent manner rather than the frame-shaped manual design of the input-independent prompt (or program) of VP (or BAR). (2) While VP relies on first-order algorithms and BAR adopts the RGF, we utilize the new variants of SPSA \cite{spall1998overview}, SPSA-GC, which is enhanced with a proper modification in the parameter update rule. (3) Contrary to the medical imaging-only validation in BAR, based on the above two technical difference, BlackVIP successfully adapt the pre-trained model to diverse data domains (described in Section B.1.).

\subsection{Implementation Details}
\label{app:b_implementation}
\paragraph{Architecture}
For the fixed text prompt design of each dataset, those are shared across all baseline methods and BlackVIPs, we use the same templates provided by  \cite{bahng2022visual} for SVHN, CLEVR, and Resisc45, and \cite{zhou2022learning} for the remaining 11 datasets. For the fixed encoder part of our \textit{Coordinator}, we use the ImageNet pre-trained \texttt{vit-mae-base} checkpoint\footnote{\url{https://huggingface.co/docs/transformers/model_doc/vit_mae}} from HuggingFace. The output shape of the encoder is $N\times768$, where $N$ is the number of instances in the batch. For the BlackVIP-SE, we used scikit-learn's vanilla PCA or kernel PCA method\footnote{We considered whether to use a kernel or not and which kernel (among RBF, cosine, and polynomial kernels) to use as hyperparameters.}. For fast training, we pre-computed the PCA features for BlackVIP-SE and retrieved them by index during each training iteration. During the inference phase, we load the trainset-fitted PCA projection matrix and apply it to the flattened test input $x$ to get the feature vector fed into the prompt decoder. Here, we design the decoder based on \textit{depth-wise separable convolution} (DSC) layer \cite{chollet2017xception} for parameter efficiency. Specifically, we build a block of [\texttt{NORM-ACT-CONV}] and stack it five times. The \texttt{NORM} and \texttt{ACT} denote Batch Normalization and Gaussian Error Linear Unit, respectively. The \texttt{CONV} operation of the first four blocks is DSC, and the last one is a standard convolutional layer. Our implementation code is available at \url{https://github.com/changdaeoh/BlackVIP}

To satisfy a fully convolutional design without loss of expressiveness, tensors that are fed into the decoder must be shaped in a 3D feature map. For this, we additionally govern a task-specific single continuous vector $\phi_{t}$ (called \textit{prompt trigger vector}), which is concatenated with the output feature vector of encoder, leading to the appropriate size of 1D vector for reshaping to a 3D tensor. For the BlackVIP, we set the dimension of the prompt trigger vector to 800, resulting in 1568 dimensions of a concatenated vector that can be reshaped to $32\times7\times7$ shaped 3D tensor. Meanwhile, BlackVIP-SE allows users to set the latent feature dimension (for the reduced PCA coordinate system) according to their own preference. This contributes to further reducing the input dimension of the decoder feature map, and we set the PCA feature dimension to 98. Rather than concatenation, we simply use the sum of the trigger vector and PCA feature as input to the prompt decoder. The prompt trigger is shared across all instances for a given task.

\paragraph{Optimization and other configurations}
For a stable approximation of the gradient in practice, ZOO algorithms repeat the gradient estimation step for several times and use the mean of those estimates as a final approximation of the gradient. Usually, the approximation quality is proportional to the number of these repeats. We set this repeat as five times for all baselines that use ZOO. 

Besides the learning rate and learning rate schedule parameters, ZOO algorithms have some additional algorithm-specific hyperparameters that need to be tuned. For RGF, these are the standard deviation of a random Gaussian vector and a smoothing parameter, and for SPSA, these are the perturbation magnitude and its decaying factor. We provide the search range of each hyperparameter in Table \ref{tab:a_hyparam}. The search range for algorithm-specific parameters is based on the proposal of authors of SPSA \cite{119632} and BAR \cite{tsai2020transfer}. Moreover, among the valid perturbation distributions of SPSA, we adopt the Segmented Uniform $[-1.0, -0.5] \cup [0.5, 1.0]$.

The learning objective is a cross-entropy loss for VP and BlackVIP and a focal loss for BAR (following \cite{tsai2020transfer}). For all black-box approaches, the batch size is set to 128 across all datasets. Except for the SUN397 (1,000), StanfordCars (2,500), and ImageNet (500), we optimize all methods during 5,000 epochs for convergence. Note that the input space visual prompting with first-order algorithm already requires sufficiently large iterations, e.g., 1,000 epochs \cite{bahng2022visual} with full dataset, and ZOO demands many more iterations due to the lack of gradient information. 

\subsection{Hyperparameter Sweep}
\label{app:b_hyperparameter}
In this section, we provide the hyperparameter search range of each algorithm, summarized in Table \ref{tab:a_hyparam}.

\begin{table}[htbp]
\centering
\caption{Hyperparameter sweep. Large LR (learning rate) of BAR and VP is based on \cite{bahng2022visual} to directly optimize pixel values rather than the neural network's weights. PM denotes perturbation scale $c_{i}$, and parameters of the dimensionality reduction algorithm.}
\label{tab:a_hyparam}
\resizebox{0.5\textwidth}{!}{\begin{tabular}{@{}lcc@{}}
\toprule
Hyperparameter & Algorithm & Search Range \\ \midrule
initial LR & BAR, VP & $\{$40.0, 20.0, 10.0, 5.0, 1.0$\}$ \\
initial LR ($a_{1}$) & BlackVIPs & $\{$1.0, 0.1, 0.01, 0.005$\}$ \\
min LR & BAR & $\{$0.1, 0.01, 0.001$\}$ \\
decaying step & BAR & $\{$0.9, 0.5, 0.1$\}$ \\
LR decaying factor & VP, BlackVIPs & $\{$0.6, 0.5, 0.4, 0.3$\}$ \\
initial PM ($c_{1}$) & BlackVIPs & $\{$0.01, 0.005, 0.001$\}$ \\
PM decaying factor & BlackVIPs & $\{$0.2, 0.1$\}$ \\
std. of perturbation & BAR & $\{$1.0, 0.5$\}$ \\
smoothing & BAR & $\{$0.1, 0.01, 0.001$\}$ \\ 
gradient smoothing & VP, BlackVIPs & $\{$0.9, 0.7, 0.5, 0.3$\}$ \\
dimension of features & BlackVIP-SE & $\{$32, 49, 98, 196$\}$ \\ 
PCA kernel type & BlackVIP-SE & $\{$linear, RBF, cosine, polynomial$\}$ \\ \bottomrule
\end{tabular}}
\end{table}

\section{Detail Description of BlackVIP's \textit{Coordinator}}
\label{app:c_model_desc}
On the transfer learning of a pre-trained model which provides no accessibility about any architectural information or actual model parameters, BlackVIP treats this situation with two novel mechanisms: (1) parameter-efficient instance-aware prompt generation network, and (2) stable zeroth-order optimization algorithm that is based on SPSA \cite{119632}. In this section, we provide a detailed description of the first component, Coordinator.

Different from existing works on visual prompting, we reparameterize the input space visual prompt $\phi$ as a neural network, \textit{Coordinator} $v_{\bm\phi}(\cdot)$ that generates an input-dependent visual prompt $v_{\bm\phi}(x)$. Coordinator is composed with encoder $f(\cdot)$, decoder $g_{\phi_{d}}(\cdot)$ and task-specific learnable vector $\phi_{t}$. The encoder is used for extracting instance-specific latent feature vector $z_{x}=f(x)$ contributing to the construction of the optimal input space visual prompt for each instance. Because our goal in this work is the broad utilization of pre-trained models on diverse downstream tasks, we adopt a pre-trained encoder network optimized by a self-supervised learning objective, not by a supervised learning objective or scratch network. Specifically, we use the ViT-B/16 weights from the \textit{Masked AutoEncoding} pre-training \cite{he2022masked}. We present the grounds for using the self-supervised learning encoder in the main paper, refer to Sec. 3. During the training phase, this pre-trained encoder part is frozen (not updated) and just acts as a feature extractor. Then, the instance-specific feature vector from the encoder is conveyed to the decoder for a prompt generation. 

Prompt decoder $g_{\phi_{d}}(\cdot)$ is a lightweight convolutional neural network, which has learnable parameters \textbf{less than 10K} by default. Note that the generated prompt has the same shape as the input image, so our prompt covers the entire region of the image, unlike previous visual prompting and reprogramming works applied to the partial region of the image.

In addition to the feature vector from the fixed encoder, the decoder also incorporates an additional input which is shared for all instances across the current dataset. The so-called \textit{prompt trigger vector} $\phi_{t}$ is a continuous vector that also contributes to the design of a visual prompt by collaborating with the instance-specific rich feature vector from the encoder. By introducing this prompt trigger vector, the decoder of the Coordinator can enjoy additional information to generate more proper prompts for a given task. Besides, it helps to build the 3D feature map for the decoder's input, which is necessary for designing a parameter-efficient fully convolutional decoder network.

\section{Additional Experimental Results}
\subsection{Further Analysis on Synthetic Dataset} 
\begin{figure}[t] \vspace{-0.75em}
     \centering
        \includegraphics[width=0.485\linewidth]{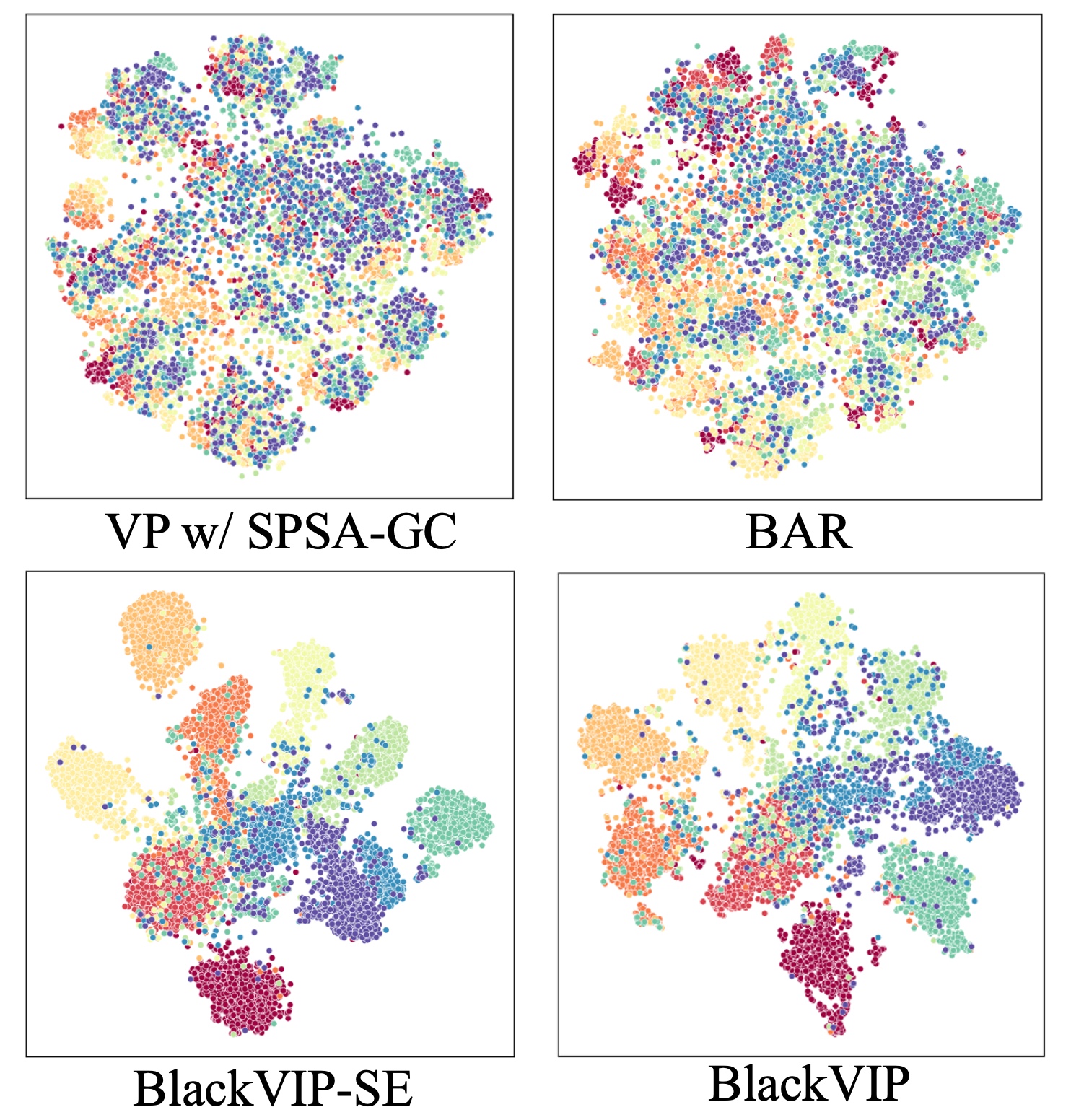}
        \includegraphics[width=0.5\linewidth]{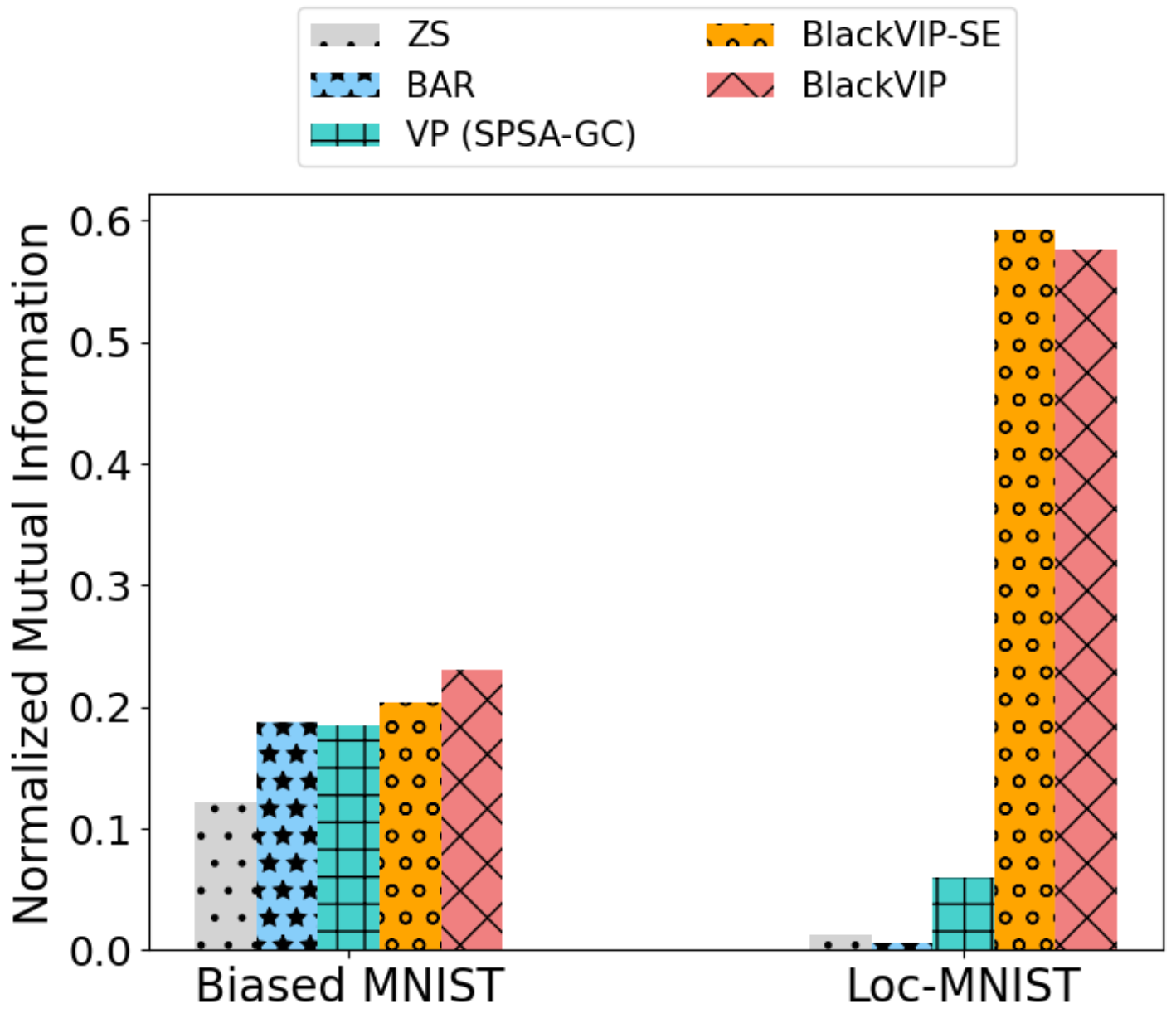}
    \caption{(Left) Embedding visualization with t-SNE \cite{van2008visualizing} on the prompt-augmented images from test splits of Loc-MNIST 1:4 setup. (right) Normalized Mutual Information (NMI) \cite{loedeman2022prompt} score of the learned embedding.}
    \label{fig:synthetic_quantative} \vspace{-0.75em}
\end{figure}

We validated our novel prompt design strategy on some challenging generalization setups, i.e., correlation shift (Biased MNIST) and varying object location (Loc-MNIST). As supporting evidence for the improved performance on these sets, we provide the visualization of the t-SNE embedding and normalized mutual information between the visual embeddings of the final layer and the ground truth labels in Fig. \ref{fig:synthetic_quantative}, where we can see that BlackVIP and BlackVIP-SE successfully discriminate different classes of images in the input space, resulting in better discriminative latent embeddings.

\subsection{Ablation Study} \label{sec:ablation}
\subsubsection{Model Architectures}
To study the versatility of our method, we vary the backbone architecture of the pre-trained target model and the encoder of Coordinator in Tab. \ref{tab:ablation_backbone}. While BAR and the naive application of SPSA-GC on VP fail to improve the zero-shot performance of CNN-based target backbones that lack the global attention of Transformers \cite{vaswani2017attention}, our BlackVIPs consistently bring huge performance gains across all the architectures. It implies that BlackVIPs are \textit{architecture-agnostic} approaches, which pursue the general adaptation method for high-performing PTMs. Regarding the comparison between BlackVIP and BlackVIP-SE, BlackVIP shows better results when the model architecture of the Coordinator encoder aligns with the target PTM's architecture. Meanwhile, BlackVIP-SE achieves more stable performance improvement across all architectures.

\begin{table}[t!]
\centering
\footnotesize
\caption{Ablation study for backbone architecture. Classification accuracy on EuroSAT across pre-trained target backbone architectures and BlackVIP's Coordinators.}
\begin{tabular}{@{}lcccc|c@{}}
\toprule
\textbf{Method} & \multicolumn{5}{c}{\textbf{Target Backbone}} \\ 
\multicolumn{1}{l}{} & RN50 & RN101 & ViT-B/32 & ViT-B/16 & \textit{Avg.} \\ \midrule
{ZS} & 37.5 & 32.6 & 45.2 & 40.8 & 48.4 \\
{BAR} &  26.9 & 33.5 & {{70.3}} & {\textbf{77.2}} & 52.0 \\
{VP w\ SPSA-GC} & {34.7}& {31.2}& {\textbf{71.1}}& {70.9} & 52.0 \\
BlackVIP-SE & 50.0 & \textbf{55.1} & 66.2 & 71.2 & \textbf{60.6} \\
BlackVIP (RN50) & {\textbf{51.3}} & {{50.8}} & {62.9} & {68.5} & {58.4} \\
BlackVIP (ViT-B/16) & {{48.4}} & {{51.3}}& {67.9} & {{73.1}} & {60.2} \\ \bottomrule
\end{tabular} \label{tab:ablation_backbone}
\end{table}
\begin{table}[htbp]
\centering
\caption{Different Coordinator weights with SPSA variants. Mean classification accuracy of three repeated runs on EuroSAT.}
\small
\begin{tabular}{@{}c|cc@{}}
\toprule
Encoder Type & Optim. & Acc. \\ \midrule
\multicolumn{2}{c}{Zero-Shot} & 47.9 \\ \midrule
\textit{scratch} & SPSA & 49.6 \\
\textit{scratch} & SPSA-GC & 49.5 \\
\textit{Sup. pre-trained} & SPSA & 59.4 \\
\textit{Sup. pre-trained} & SPSA-GC & 65.2 \\
\textit{SSL pre-tained} & SPSA & 69.4 \\ \midrule
\multicolumn{2}{l}{\textbf{BlackVIP-SE} (\textit{PCA matrix} with SPSA-GC)} & \textbf{71.2} \\
\multicolumn{2}{l}{\textbf{BlackVIP} (\textit{SSL pre-trained} with SPSA-GC)} & \textbf{73.1} \\ \bottomrule
\end{tabular} \label{tab:ablation-optim}
\end{table}

\subsubsection{Coordinator and Zeroth-order Methods} BlackVIPs adopt the encoder-decoder structure to efficiently generate the input-dependent prompts. We exploit an SSL pre-trained encoder (in BlackVIP) and PCA projection encoder (in BlackVIP-SE) while learning a randomly initialized lightweight decoder from scratch. From the design philosophy of BlackVIPs, we expect that the encoder extracts the essential features of the given image, including the spatial features, and the decoder utilizes the features to produce a spatially and semantically structured prompt tailored to the input. We conjecture that SSL pre-trained and PCA encoders are desirable to capture the demanding semantics instead of a supervised one learned from pre-defined labels. Here, Table \ref{tab:ablation-optim} confirms that the SSL encoder outperforms the supervised pre-trained or randomly initialized encoder (scratch). Besides, the PCA encoder also outperforms the large-scale supervised encoder, which indicates the potential of the classic statistical methods for feature extraction during transfer learning. Furthermore, SPSA-GC improves the 3.7\% accuracy than SPSA, from 69.4 to 73.1. It denotes that approximated gradients by our SPSA-GC are more accurate than the original SPSA.

\subsection{Query Efficiency}
\begin{figure*}
    \includegraphics[width=\textwidth]{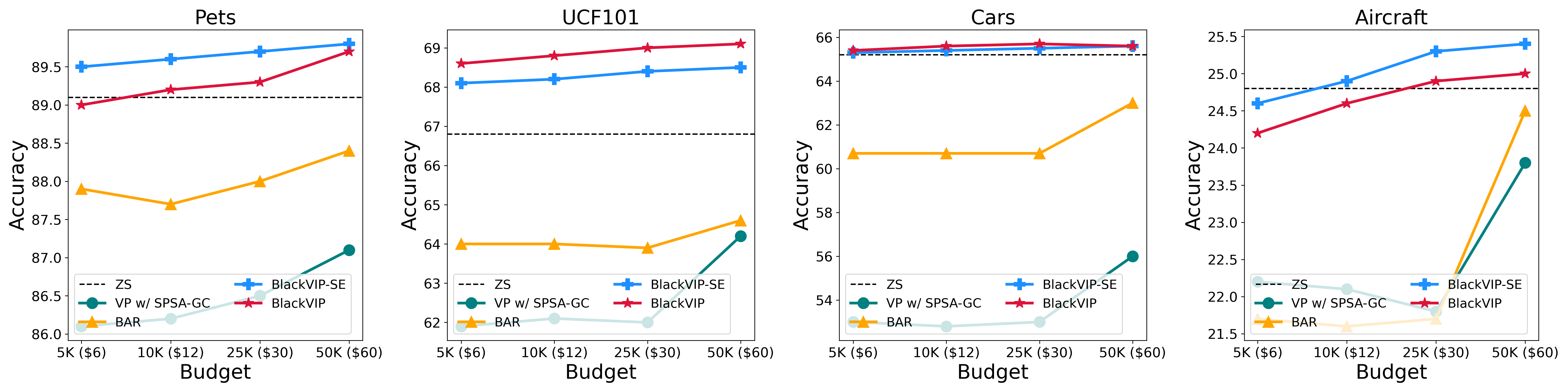}
    \caption{Classification accuracy for given queries and corresponding budget (\$ USD) of different black-box visual prompting method.}
	\label{fig:a_gcam_clevr}
\end{figure*}
We present an expanded analysis of the query efficiency experiment in the manuscript on four model datasets covering the natural object recognition (Pets), action recognition (UCF101), and fine-granular visual recognition (Cars and Aircraft). We observe outstanding query efficiency of BlackVIP and BlackVIP-SE compared with other black-box visual prompting methods, which commonly failed to achieve meaningful improvement from zero-shot inference under a few queries.

\subsection{Black-box Prompting on Larger Backbone} 
\begin{table}[!ht]
\centering
\caption{Results on black-box visual prompting with CLIP VIT-L/14 backbone on three datasets (reported in Accuracy).}
    \begin{tabular}{ll|ccc}
    \toprule
        Backbone & Method & SVHN & CLEVR & EuroSAT \\ \midrule
        \multirow{3}{*}{CLIP ViT-B/16} & ZS & 18.1 & 14.5 & 47.9 \\ 
         & BAR & 34.9 & 18.7 & \textbf{77.2} \\
         & BlackVIP & \textbf{44.3} &\textbf{ 36.8} & 73.1 \\ \midrule
        \multirow{3}{*}{CLIP ViT-L/14} & ZS & 36.3 & 17.4 & 58.1 \\ 
         & BAR & 40.8 & 31.0 & 76.2 \\ 
         & BlackVIP &\textbf{ 54.7} & \textbf{37.0} & \textbf{80.4} \\ \bottomrule
    \end{tabular} \label{tab:apdx:larger_backbone}
\end{table}
In the main body of the paper, we focused our validation mainly on the CLIP ViT-B/16 model as our target PTM. Here, in Table~\ref{tab:apdx:larger_backbone}, we additionally report the results on ViT-L/14 on three datasets: SVHN, CLEVR, and EuroSAT, and see consistent performance gain compared to the competative baseline BAR.

\subsection{Visual Prompting in the Wilds}
\begin{table}[!t]
    \centering
    \caption{\cd{Robustness comparison under real-world distribution shift.}}
    \begin{tabular}{@{}l|cccc@{}}
    \toprule
        Method & iWildCam & FMoW & Camelyon17 & \textit{Avg.} \\ \midrule
        ZS & 9.9 & 18.9 & 55.2 & 28.0 \\ 
        BAR & 11.0 & 22.4 & 76.8 & 36.7 \\ 
        VP w/ SPSA-GC & 10.9 & 22.3 & 66.5 & 32.5 \\ 
        BlackVIP-SE & 10.6 & 21.9 & \textbf{79.7} & \textbf{37.4} \\ 
        BlackVIP & \textbf{12.3} & \textbf{22.6} & 76.5  & {37.1} \\ \bottomrule
    \end{tabular} \label{tab:wilds}
\end{table}

\cd{
These days, our AI models commonly encounter distribution shifts in real-world applications. In the manuscript, we evaluate the robustness of prompting-based adaptation approaches against synthetic correlation shift environments.
We now validate those approaches on three kinds of real-world datasets (iWildCam, FMoW, and Camelyon17) from the WILDS distribution shift benchmark \cite{koh2021wilds}. In iWildCam, the task is to classify the 182 species of animals in the wild under different remote sensing conditions (location of camera traps). In FMoW, satellite images should be classified into the building and land types, 62 in total, and the shifting factor is the combination of (year, region). Meanwhile, Camelyon17 requires a binary classification (tumor or normal) on the patch-level tissue slides under varying hospital imaging conditions. 
In Table \ref{tab:wilds}, the baseline approaches, BAR and VP w/ SPSA-GC, show their effectiveness in improving the zero-shot classifier's performance, and our BlackVIPs achieve superior average performances. Interestingly, while BlackVIP-SE underperforms other methods in iWildCam and FMoW, it shows outstanding performance on Camelyon17, where we speculate the intrinsic dimensionality of the task is relatively low, and the PCA features can be discriminative enough to generate good prompts. 
}
\subsection{Visual Prompting for Quantized Models}
\begin{table}[!ht] \vspace{-0.5em}
    \centering
    \caption{Image classification with quantized models (CLIP ViT-B/16). Evaluation is conducted on CPU.}
    \scriptsize
    \begin{tabular}{@{}cl|cc|c@{}}
    \toprule
        Quantization & Method & SVHN & CLEVR & Inference time \\ \midrule
        \multirow{3}{*}{16-bit, float} & ZS & 18.0 & 14.7 & 0.3177 \\ 
        & VP w/ SPSA-GC & 34.8 & 21.8 & 0.3236 \\ 
        & BlackVIP-SE & \textbf{47.2} & 29.3 & 0.3223 \\ 
        & BlackVIP & 43.8 & \textbf{34.9} & 0.3489 \\ \midrule
        \multirow{3}{*}{8-bit, int} & ZS & 15.0 & 16.6 & 0.0482 \\ 
        & VP w/ SPSA-GC & 14.7 & 14.2 & 0.0519 \\ 
        & BlackVIP-SE & \textbf{16.9} & 15.4 & 0.0548 \\ 
        & BlackVIP & 16.3 & \textbf{18.7} & 0.0584 \\ \bottomrule
    \end{tabular} \label{tab:quantization} \vspace{-1.5em}
\end{table}
While the main goal of BlackVIPs is to enable memory-efficient transfer learning without parameter accessibility, as the scale of modern state-of-the-art AI models continually skyrockets \cite{kaplan2020scaling}, reducing the inference latency is also increasingly crucial for deploying those models on diverse edge devices. Post-training quantization is a representative remedy that reduces the model size and required peak memory by converting the learned model parameters into low-bit ones. Here, we validate the visual prompting strategies on post-training quantized low-bit models. Specifically, we produce the visual prompts learned by the original 32-bit model and equip the prompts with images to feed into the converted low-bit model for performance evaluation. While the effectiveness varies across the bit-size and dataset, we confirm that (in Table \ref{tab:quantization}) BlackVIPs can actually collaborate with the post-training quantization method to reduce the memory requirement. Compared with VP w/ SPSA-GC, BlackVIPs show remarkably better performance while not significantly increasing inference-time latency.

\subsection{Grad-CAM Analysis}
\label{app:a_qual_anal}
\begin{figure*} 
    \centerline{\includegraphics[width=0.76\textwidth]{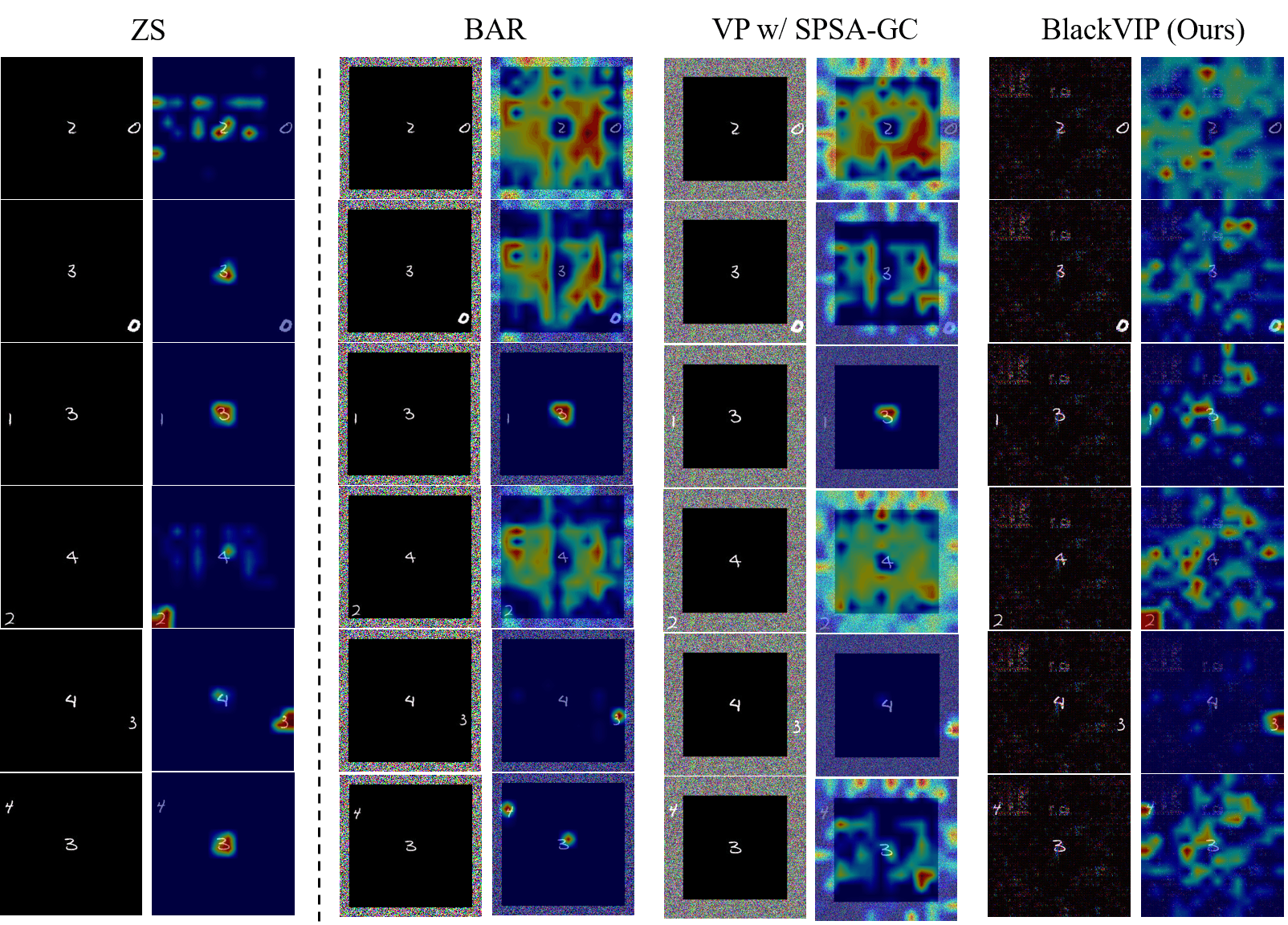}}
    \caption{Grad-CAM on Loc-MNIST. Compared to baseline methods, BlackVIP effectively adapts the model to aim at edge-located true digit corresponding true label rather than the obstructive fake digit in the center of the image.}
	\label{fig:a_gcam_lmnist}
\end{figure*}

To investigate whether visual prompts produced by each method adapt the pre-trained model, we visualize the Grad-CAM \cite{selvaraju2017grad} on the original image and prompted image of the encoder's penultimate layer (Figure \ref{fig:a_gcam_clevr}-\ref{fig:a_gcam_lmnist}). We select eight datasets that represent the diverse image domains and experimental conditions: (\textit{Natural}) OxfordPets and SVHN, (\textit{Specialized}) EuroSAT, (\textit{Structured}) CLEVR-count, (\textit{Action Recognition}) UCF101, (\textit{Fine-Grained}) StanfordCars, (\textit{Synthetic}) Biased-MNIST and Loc-MNIST. Detail descriptions for each dataset are provided in Sec. A.1.

\begin{figure*} 
    \centerline{\includegraphics[width=0.76\textwidth]{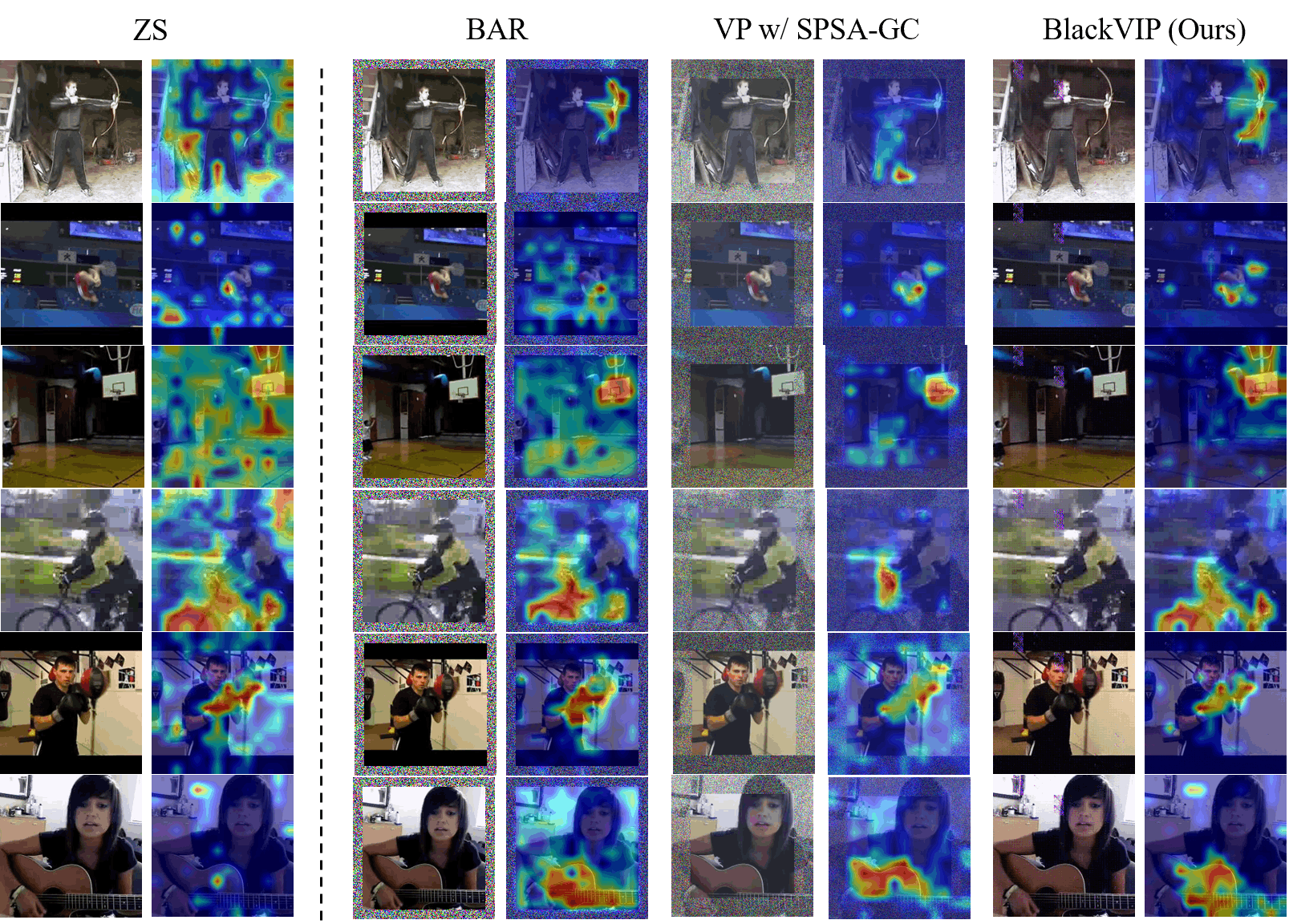}}
    \caption{Grad-CAM on UCF101. Compared to baseline methods, BlackVIP concentrates the attention of models on local areas of the image for effective recognition of the specific actions.}
	\label{fig:a_gcam_ucf}
\end{figure*}

BlackVIP generates diverse prompts to properly adapt the pre-trained model to the targeted data domains. When the task is counting the number of objects (CLEVR) in the entire region of an image, BlackVIP extends the attention of the pre-trained model to the whole view of an image as shown in Figure \ref{fig:a_gcam_clevr}. If the task requires a fine-grained understanding of objects or recognition of actions (UCF101), BlackVIP concentrates the model's attention on class-related regions, as shown in Figure \ref{fig:a_gcam_ucf}.

\begin{figure*}
    \centerline{\includegraphics[width=0.76\textwidth]{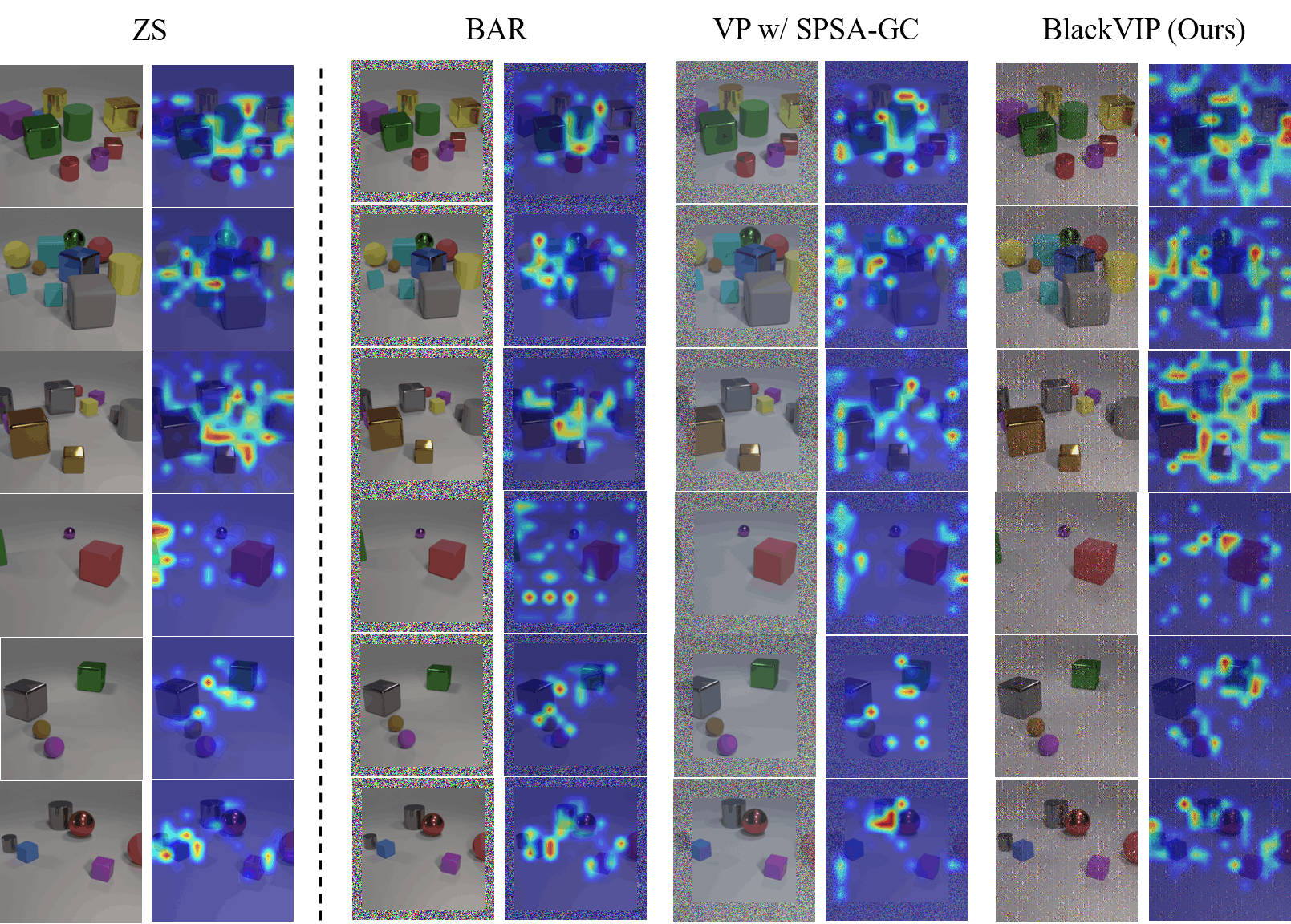}}
    \caption{Grad-CAM on CLEVR. Compared to baseline methods, BlackVIP extends the attention of models to broad areas of the image for effective reasoning on the number of objects.}
	\label{fig:a_gcam_clevr}
\end{figure*}
\begin{figure*} 
    \centerline{\includegraphics[width=0.75\textwidth]{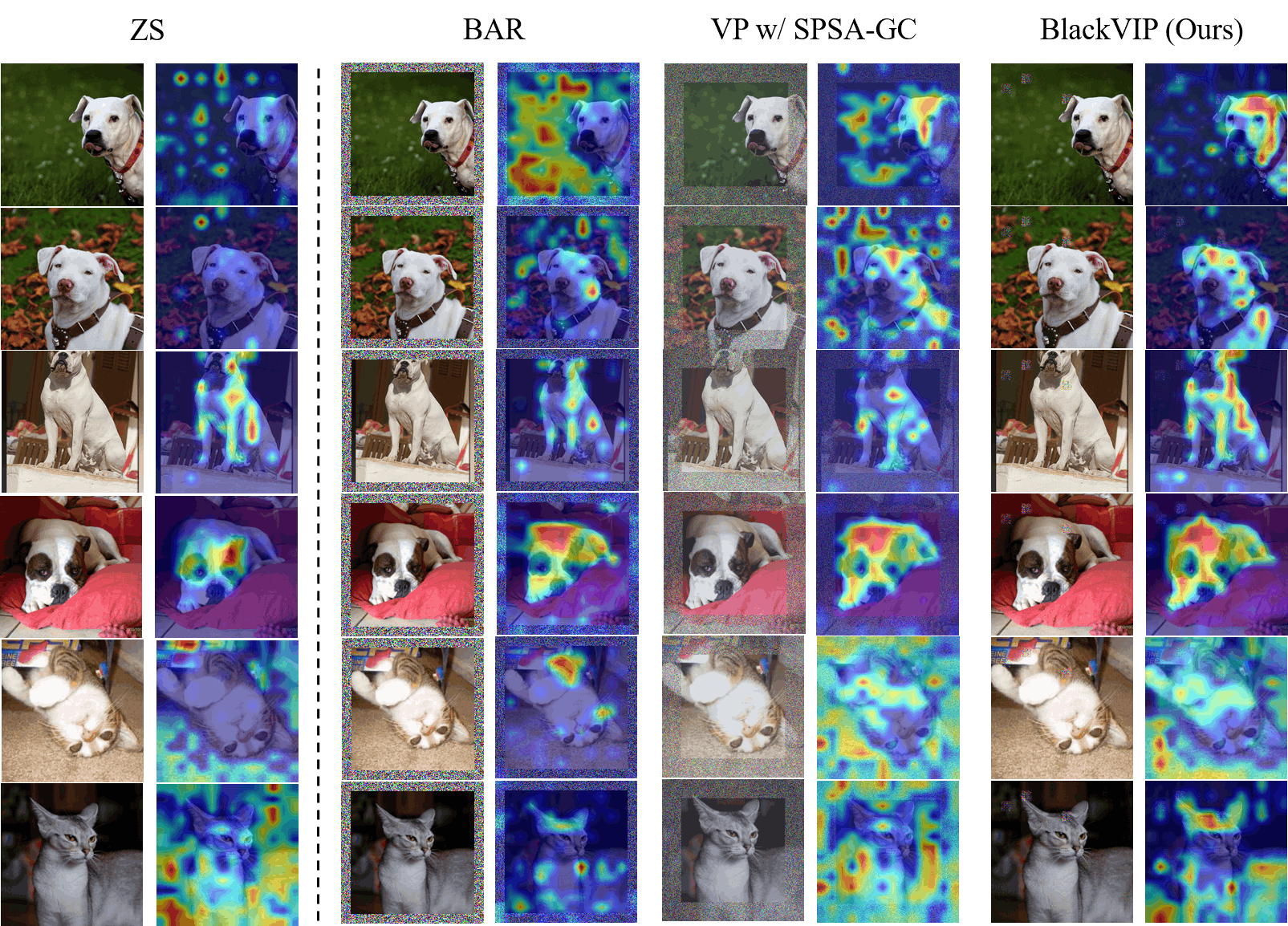}}
    \caption{Grad-CAM on OxfordPets. Compared to baseline methods, BlackVIP effectively adapts the model to focus on the target object rather than spurious features such as the background.}
	\label{fig:a_gcam_pets}
\end{figure*}
\begin{figure*} 
    \centerline{\includegraphics[width=0.76\textwidth]{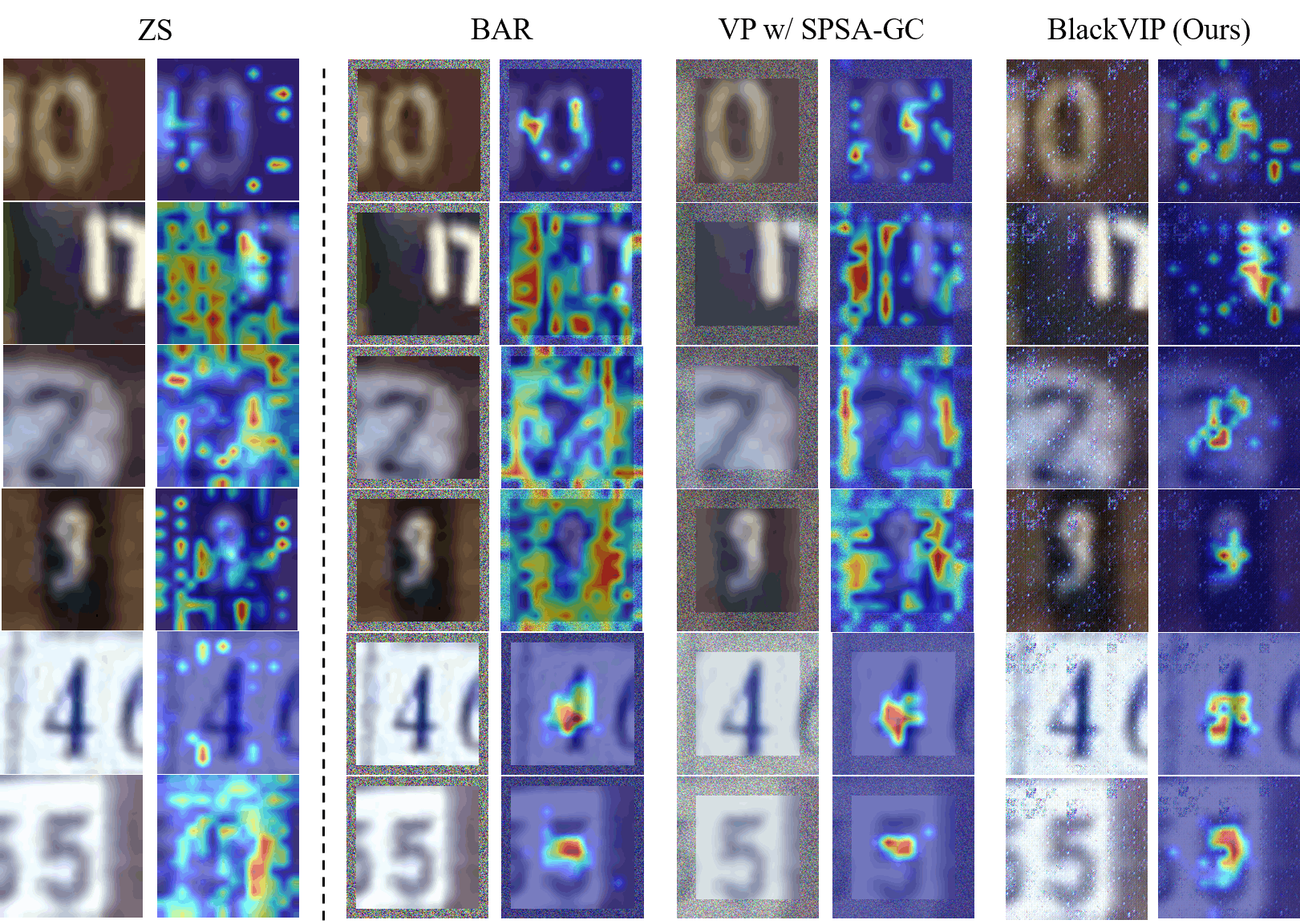}}
    \caption{Grad-CAM on SVHN. Compared to baseline methods, BlackVIP effectively adapts the model to focus on the target digit rather than spurious features such as the background.}
	\label{fig:a_gcam_svhn}
\end{figure*}
\begin{figure*} 
    \centerline{\includegraphics[width=0.76\textwidth]{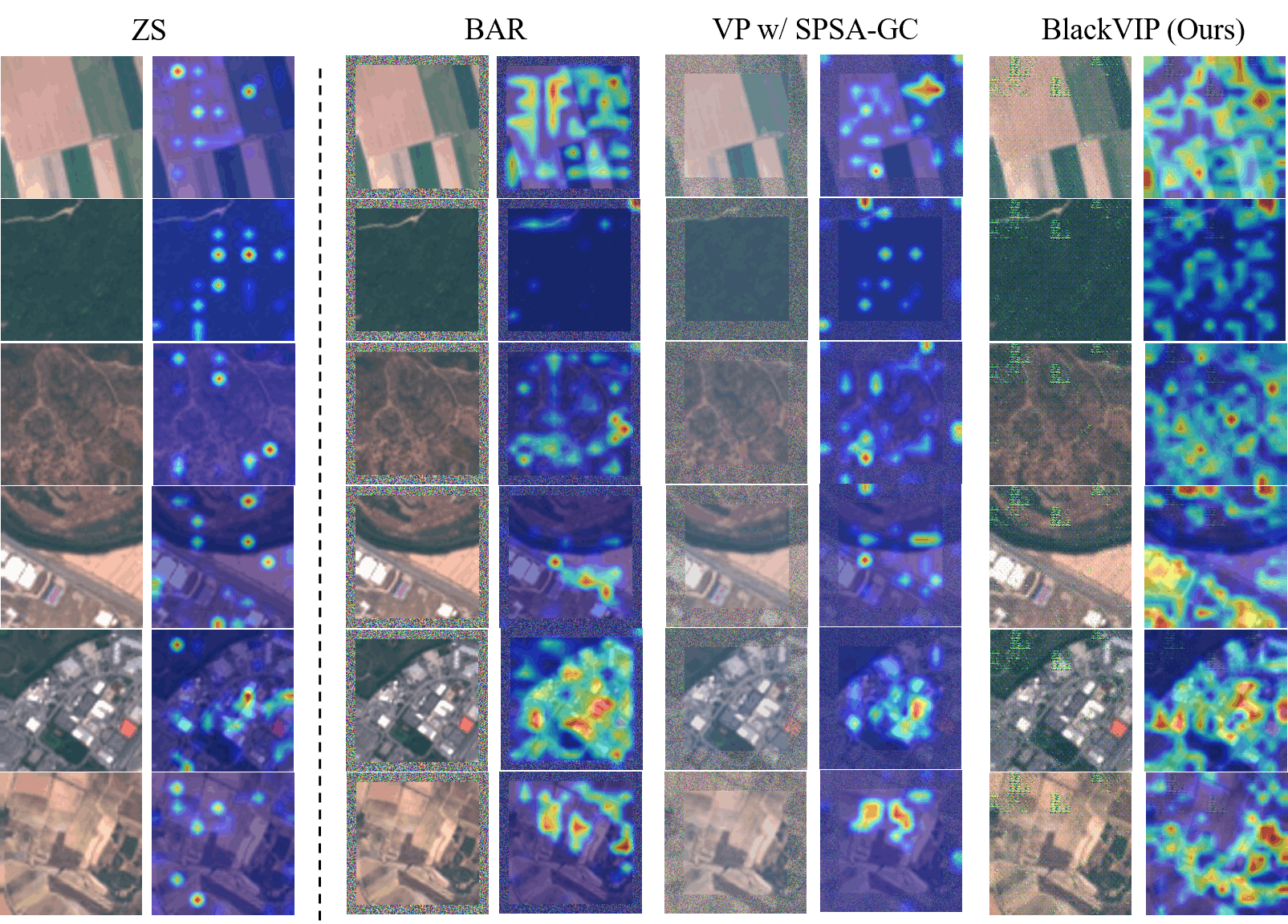}}
    \caption{Grad-CAM on EuroSAT. Compared to baseline methods, BlackVIP extends the attention of models to broad areas of the image for effective classification of satellite imagery.}
	\label{fig:a_gcam_eur}
\end{figure*}
\begin{figure*} 
    \centerline{\includegraphics[width=0.76\textwidth]{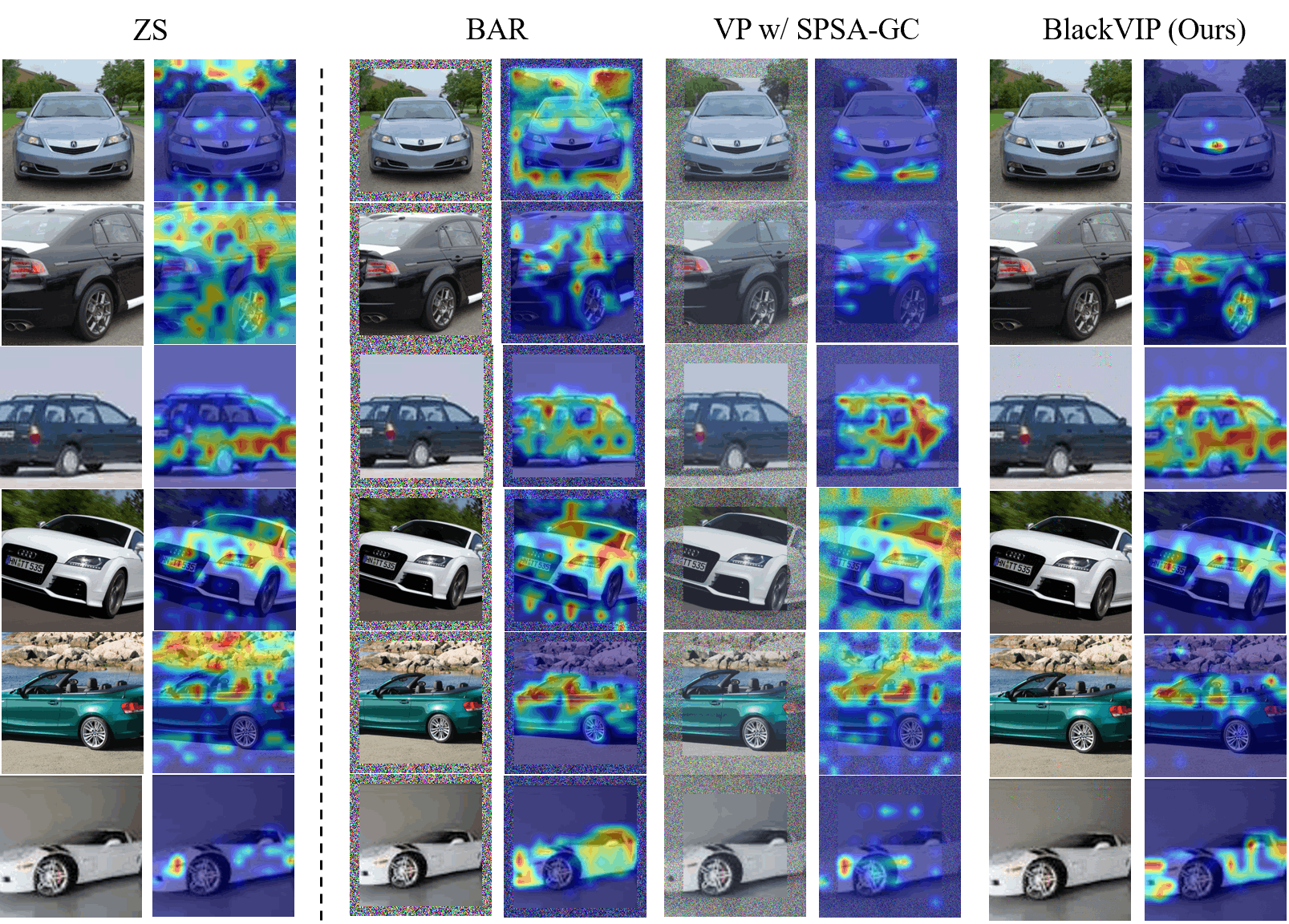}}
    \caption{Grad-CAM on StanfordCars. Compared to baseline methods, BlackVIP concentrates the attention of models on an object or local areas of an image for effective fine-grained classification.}
	\label{fig:a_gcam_cars}
\end{figure*}
\begin{figure*}
    \centerline{\includegraphics[width=0.76\textwidth]{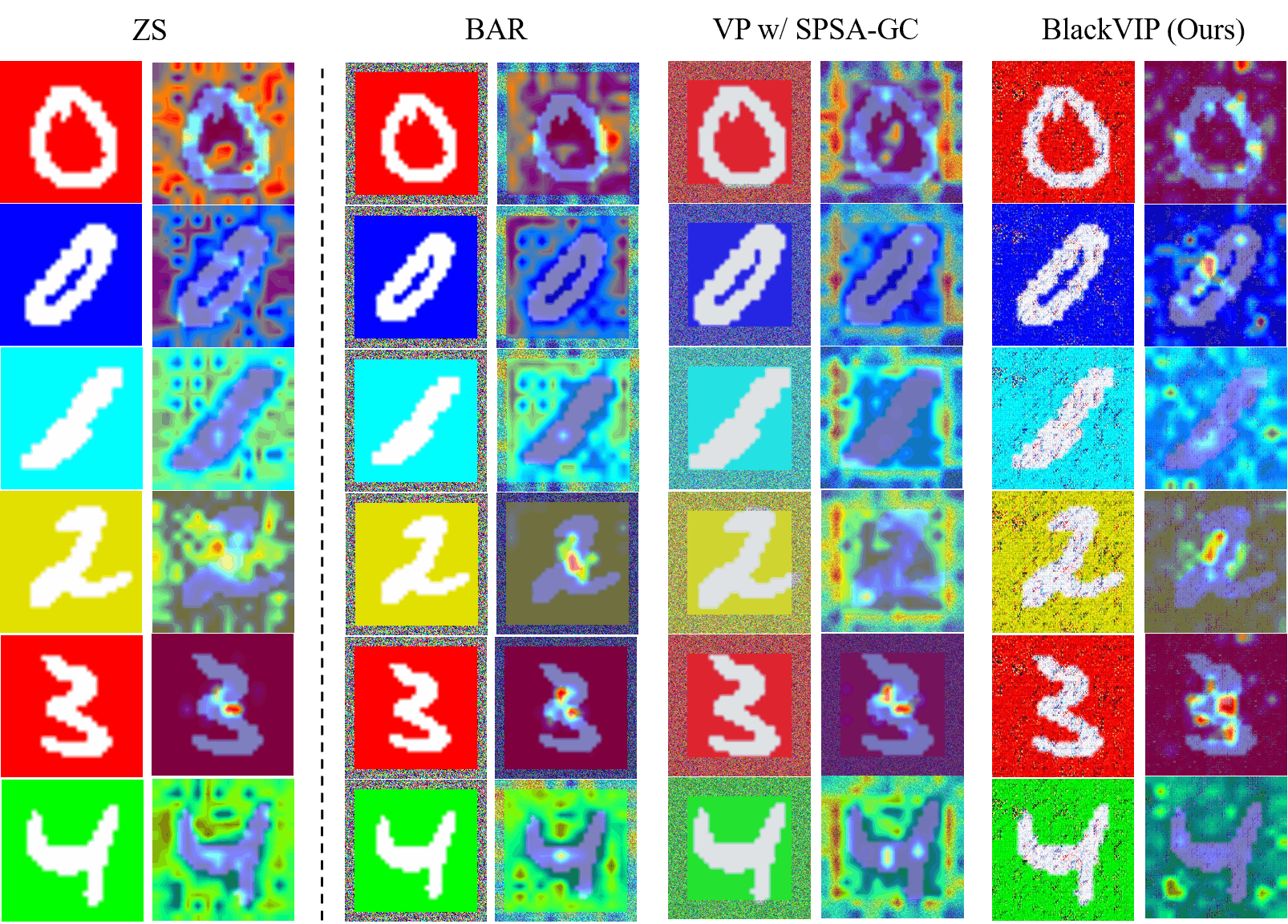}}
    \caption{Grad-CAM on Biased-MNIST. While baseline methods attend to the background rather than digit shape, our BlackVIP can bypass this spurious feature through a widely scattered visual prompt and focus more of the attention on the shape of the digit.}
	\label{fig:a_gcam_bmnist}
\end{figure*}



\end{document}